\documentclass[10pt,twocolumn,letterpaper]{article}

\usepackage{cvpr}
\usepackage{times}
\usepackage{epsfig}
\usepackage{graphbox,graphicx}
\usepackage{amsmath}
\usepackage{amssymb}
\usepackage{multirow}
\usepackage{subfigure}
\usepackage{colortbl}
\usepackage{verbatim}
\usepackage{romannum}
\usepackage{siunitx}

\definecolor{Gray}{gray}{0.9}
\newcolumntype{g}{>{\columncolor{Gray}}l}
\newcommand*\samethanks[1][\value{footnote}]{\footnotemark[#1]}

\usepackage[pagebackref=true,breaklinks=true,letterpaper=true,colorlinks,bookmarks=false]{hyperref}

\cvprfinalcopy %

\def\cvprPaperID{3142} %

\ifcvprfinal\fi
\begin{document}
\pagenumbering{arabic} 

\title{Deep High-Resolution Representation Learning for Human Pose Estimation}

\author{Ke Sun$^{1,2}$$\thanks{Equal contribution.}$ $\thanks{This work is done when Ke Sun was an intern at Microsoft Research, Beijing, P.R. China}$   ~~~ Bin Xiao$^{2}$\samethanks[1] ~~~ Dong Liu$^{1}$ ~~~ Jingdong Wang$^{2}$\\
$^{1}$University of Science and Technology of China~~~ $^{2}$Microsoft Research Asia\\
{\tt\small \{sunk,dongeliu\}@ustc.edu.cn, \{Bin.Xiao,jingdw\}@microsoft.com}
}

\maketitle

\begin{abstract}
In this paper, we are interested in 
the human pose estimation problem
with a focus on learning reliable high-resolution representations.
Most existing methods
\emph{recover high-resolution representations
from low-resolution representations}
produced by
a high-to-low resolution network.
Instead,
our proposed network~\emph{maintains high-resolution representations}
through the whole process.

We start from a high-resolution subnetwork as the first stage,
gradually add high-to-low resolution subnetworks one by one
to form more stages,
and connect
the mutli-resolution subnetworks \emph{in parallel}.
We conduct \emph{repeated multi-scale fusions}
such that each of the high-to-low resolution representations
receives information from other parallel representations
over and over,
leading to rich high-resolution representations.
As a result,
the predicted keypoint heatmap 
is potentially more accurate and spatially more precise.
We empirically demonstrate the effectiveness
of our network
through the superior pose estimation results
over two benchmark datasets: the COCO keypoint detection dataset and the MPII Human Pose dataset.
In addition, we show the superiority of our network in pose tracking on 
the PoseTrack dataset. 
The code and models have been publicly available at \url{https://github.com/leoxiaobin/deep-high-resolution-net.pytorch}.

\end{abstract}

\section{Introduction}
$2$D human pose estimation has been a fundamental yet 
challenging problem in computer vision.
The goal is to localize human anatomical keypoints
(e.g., elbow, wrist, etc.) or parts.
It has many applications,
including human action recognition, 
human-computer interaction, animation, etc. 
This paper is interested in single-person pose estimation,
which is the basis of other related problems,
such as multi-person pose estimation~\cite{CaoSWS17, InsafutdinovPAA16,KocabasKA18,NewellHD17, PapandreouZKTTB17,Sekii18,NieFXY18,PapandreouZCGTK18, FangXTL17, XiaWCY17}, video pose estimation and tracking~\cite{PfisterCZ15,XiaoWW18}, etc.

The recent developments show that
deep convolutional neural networks have achieved the state-of-the-art performance.
Most existing methods
pass the input through a network,
typically 
consisting of high-to-low resolution subnetworks
that are connected in series,
and then \emph{raise the resolution}.
For instance,
Hourglass~\cite{NewellYD16} 
recovers the high resolution 
through a symmetric low-to-high process.
SimpleBaseline~\cite{XiaoWW18} adopts a few 
transposed convolution layers for generating high-resolution representations.
In addition, dilated convolutions are also used
to blow up the later layers
of a high-to-low resolution network (e.g., VGGNet or ResNet)~\cite{InsafutdinovPAA16,YangLOLW17}.

\begin{figure}[t]
\small
\centering
\includegraphics[width=1.0\linewidth]{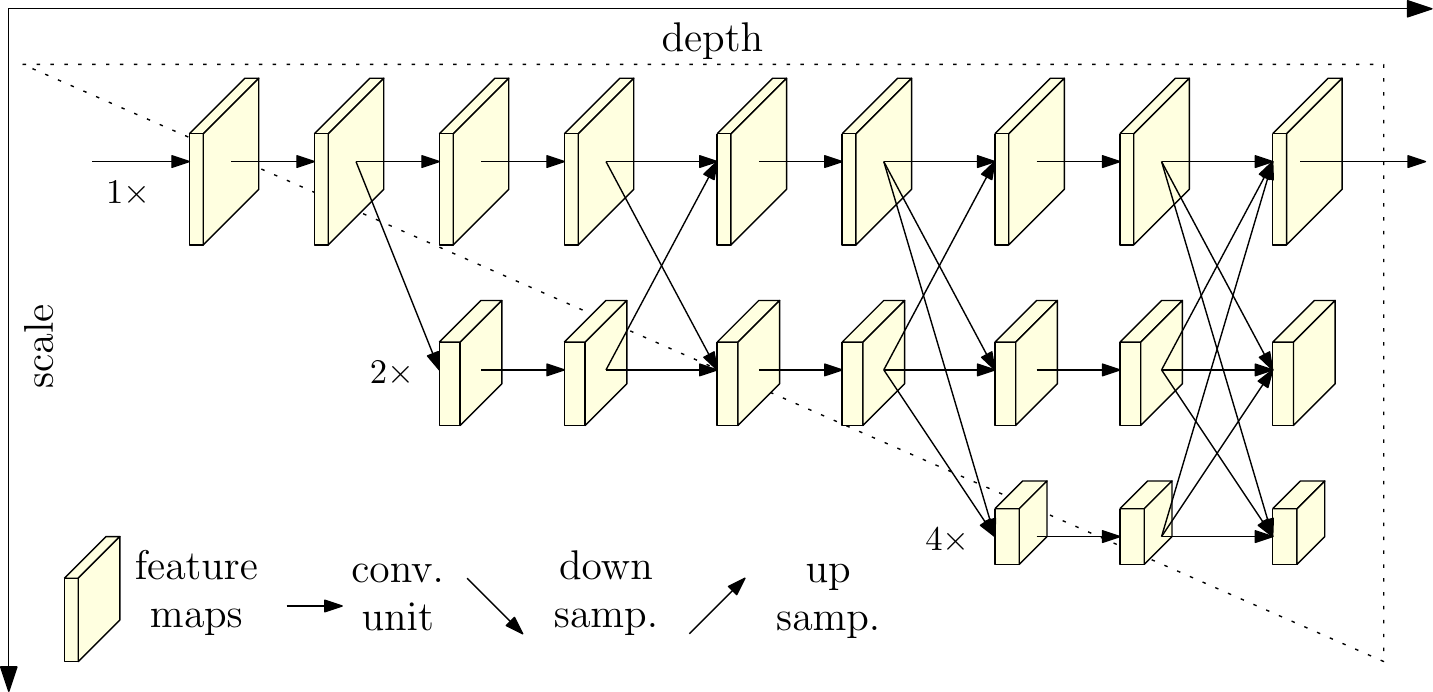}\\
\vspace{.2cm}
\caption{\small Illustrating
the architecture of the proposed HRNet.
It consists of parallel high-to-low resolution subnetworks
with repeated information exchange 
across multi-resolution subnetworks (multi-scale fusion).
The horizontal and vertical directions correspond
to the depth of the network and the scale of the feature maps, respectively.
}
\label{fig:trianglenet}
\vspace{-.5cm}
\end{figure}

We present a novel architecture,
namely High-Resolution Net (HRNet),
which is able to \emph{maintain high-resolution representations} 
through the whole process.
We start from a high-resolution subnetwork as the first stage,
gradually add high-to-low resolution subnetworks one by one
to form more stages,
and connect the multi-resolution subnetworks
in parallel.
We conduct repeated multi-scale fusions
by exchanging the information
across the parallel multi-resolution subnetworks
over and over through the whole process.
We estimate the keypoints %
over the high-resolution representations output by our network.
The resulting network is illustrated in Figure~\ref{fig:trianglenet}.

\begin{figure*}[t]
    \centering
    \small
    \includegraphics[align=t,width=.49\linewidth]{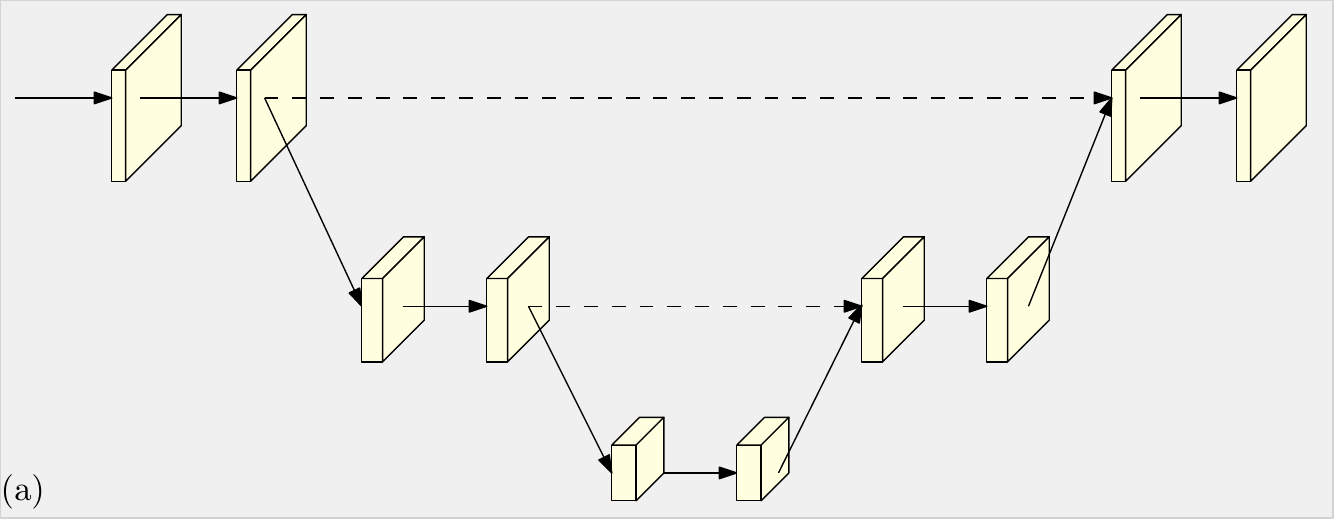}\hspace{.15cm}
    \includegraphics[align=t,width=.49\linewidth]{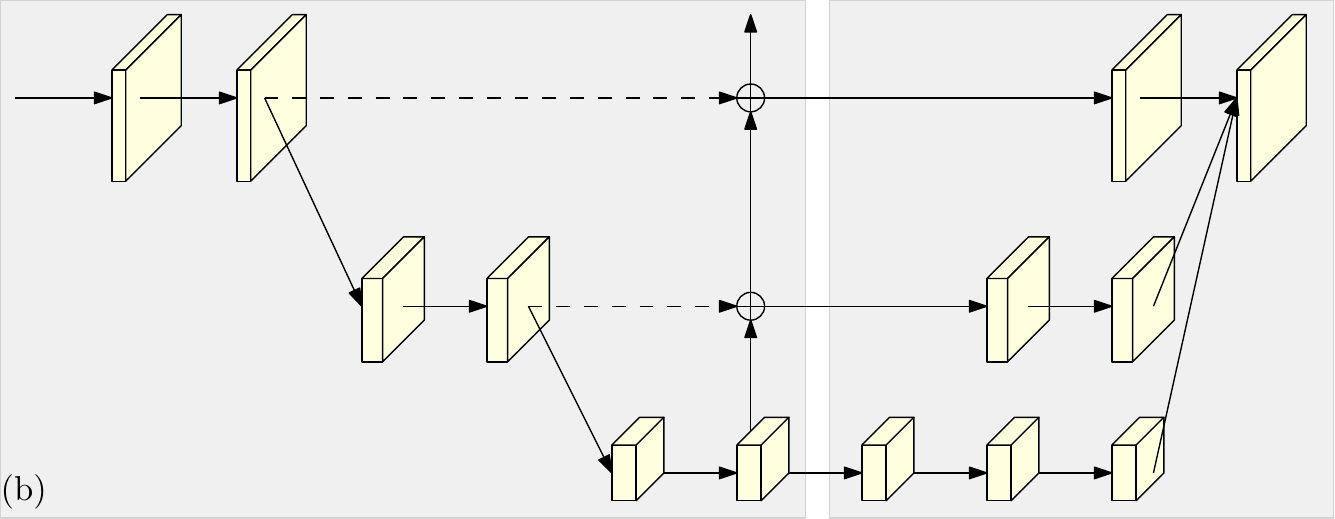}   \\ 
    \vspace{.15cm}
    \includegraphics[align=t,width=.49\linewidth]{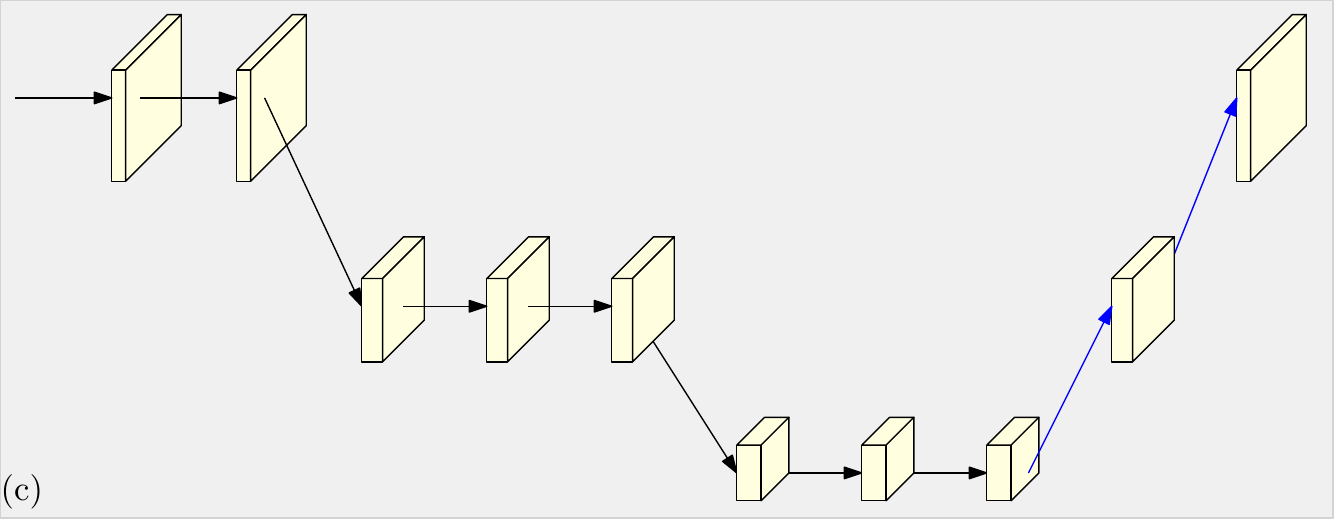}\hspace{.15cm}
    \includegraphics[align=t,width=.49\linewidth]{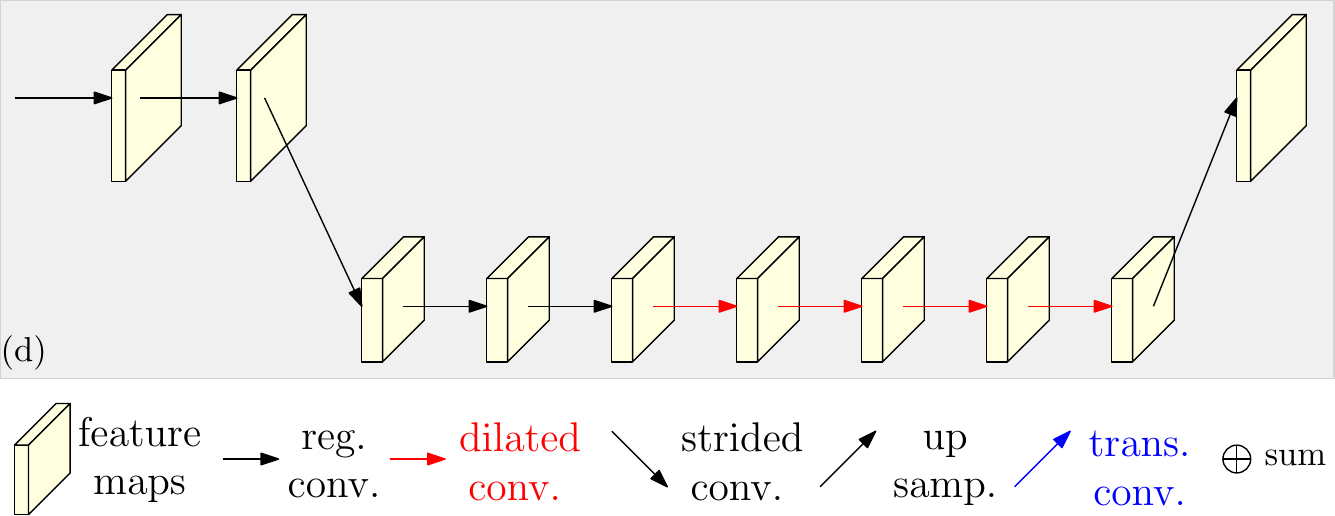}\\
    \vspace{.2cm}
    \caption{\small Illustration of representative pose estimation networks
    that rely on the high-to-low and low-to-high framework.
    (a) Hourglass~\cite{NewellYD16}. 
    (b) Cascaded pyramid networks~\cite{ChenWPZYS17}.
    (c) SimpleBaseline~\cite{XiaoWW18}: transposed convolutions for low-to-high processing.
    (d) Combination with dilated convolutions~\cite{InsafutdinovPAA16}.
    Bottom-right legend:
    reg. = regular convolution,
    dilated =  dilated convolution,
    trans. = transposed convolution,
    strided = strided convolution,
    concat. = concatenation.
    In (a), the high-to-low and low-to-high processes are symmetric.    
    In (b), (c) and (d), the high-to-low process,
    a part of a classification network (ResNet or VGGNet),  
    is \emph{heavy}, and the low-to-high process is \emph{light}.
    In (a) and (b), 
    the skip-connections (dashed lines) 
    between the same-resolution layers
    of the high-to-low and low-to-high processes
    mainly aim to fuse low-level and high-level features.
    In (b), the right part, refinenet,
    combines the low-level and high-level features 
    that are processed through convolutions.
    }
    \label{fig:summaryofposenetworks}
    \vspace{-.3cm}
\end{figure*}

Our network has two benefits 
in comparison to existing
widely-used networks~\cite{NewellYD16, InsafutdinovPAA16, YangLOLW17, XiaoWW18} for pose estimation.
(\romannum{1})
Our approach connects 
high-to-low resolution subnetworks
in parallel 
rather than in series as done in most existing solutions.
Thus, our approach is able to {maintain} the high resolution
instead of recovering the resolution through a low-to-high process,
and accordingly the predicted heatmap is potentially spatially more precise.
(\romannum{2})
Most existing fusion schemes
aggregate low-level
and high-level representations.
Instead, we perform repeated multi-scale fusions
to boost the high-resolution representations 
with the help of the low-resolution representations
of the same depth and similar level,
and vice versa,
resulting in that high-resolution representations 
are also rich for pose estimation. 
Consequently,
our predicted heatmap is potentially more accurate.

We empirically demonstrate the superior keypoint detection performance 
over two benchmark datasets: the COCO keypoint detection dataset~\cite{LinMBHPRDZ14}
and 
the MPII Human Pose dataset~\cite{AndrilukaPGS14}.
In addition,
we show the superiority of our network
in video pose tracking
on the PoseTrack dataset~\cite{andriluka2018posetrack}.

\section{Related Work}
Most traditional solutions to single-person pose estimation
adopt the probabilistic graphical model
or the pictorial structure model~\cite{YangR11,PishchulinAGS13},
which is recently improved by exploiting deep learning 
for better modeling the unary and pair-wise energies~\cite{ChenY14, TompsonJLB14, OuyangCW14} 
or imitating the iterative inference process~\cite{ChuOLW16}.
Nowadays,
deep convolutional neural network provides 
dominant solutions~\cite{GkioxariTJ16, LifshitzFU16, TangYW18,NieFY18,NieFZY18,PengTYFM18,SunLXZLW17, FanZLW15}.
There are two mainstream methods:
regressing the position of keypoints~\cite{ToshevS14,CarreiraAFM16},
and estimating keypoint heatmaps~\cite{ChuOLW16,ChuYOMYW17,YangOLW16}
followed by choosing the locations with the highest heat values
as the keypoints.

Most convolutional neural networks for 
keypoint heatmap estimation
consist of 
a stem subnetwork similar to the classification network, 
which decreases the resolution,
a main body producing the representations
with the same resolution as its input,
followed by a regressor estimating the heatmaps 
where the keypoint positions are estimated
and then transformed in the full resolution.
The main body mainly adopts the high-to-low 
and low-to-high framework,
possibly augmented with multi-scale fusion and intermediate (deep) supervision.

\vspace{.1cm}
\noindent\textbf{High-to-low and low-to-high.}
The high-to-low process
aims to
generate low-resolution and high-level representations, 
and the low-to-high process
aims to
produce high-resolution representations~\cite{BulatT16, ChenWPZYS17,HuR16, XiaoWW18, NewellYD16,TangYW18}.
Both the two processes are possibly repeated several times
for boosting the performance~\cite{YangLOLW17, NewellYD16, ChuYOMYW17}.

Representative network design patterns include: 
(\romannum{1}) {Symmetric high-to-low and low-to-high processes}.
Hourglass and its follow-ups~\cite{NewellYD16, ChuYOMYW17, YangLOLW17, KeCQL18} 
design the low-to-high process as a mirror of
the high-to-low process.
(\romannum{2}) {Heavy high-to-low and light low-to-high}.
The high-to-low process is based on the ImageNet classification network, 
e.g., ResNet adopted in~\cite{ChenWPZYS17,XiaoWW18},
and the low-to-high process is simply a few bilinear-upsampling~\cite{ChenWPZYS17} or transpose convolution~\cite{XiaoWW18} layers.
(\romannum{3}) {Combination with dilated convolutions}. 
In~\cite{InsafutdinovPAA16, PishchulinITAAG16, LifshitzFU16},
dilated convolutions are adopted in the last two
stages in the ResNet or VGGNet 
to eliminate the spatial resolution loss,
which is followed by a light low-to-high process to further increase the resolution,
avoiding  expensive computation cost for only using dilated convolutions~\cite{ChenWPZYS17, InsafutdinovPAA16, PishchulinITAAG16}.
Figure~\ref{fig:summaryofposenetworks} depicts
four representative pose estimation networks.

\vspace{.1cm}
\noindent\textbf{Multi-scale fusion.}
The straightforward way is 
to feed multi-resolution images separately into 
multiple networks and aggregate the output response maps~\cite{TompsonGJLB15}.
Hourglass~\cite{NewellYD16} and its extensions~\cite{YangLOLW17,KeCQL18} combine low-level features in the high-to-low process into 
the same-resolution high-level features in 
the low-to-high process progressively through skip connections.
In cascaded pyramid network~\cite{ChenWPZYS17},
a globalnet combines low-to-high level features in the high-to-low process progressively into the low-to-high process,
and then a refinenet combines the low-to-high level features that are processed through convolutions.
Our approach repeats multi-scale fusion, 
which is partially inspired by deep fusion
and its extensions~\cite{WangWZZ16, XieW0LHQ18, SunLLW18, ZhangQ0W17, ZhaoLMLZZTW18}.

\vspace{.1cm}
\noindent\textbf{Intermediate supervision.}
Intermediate supervision or deep supervision, early developed 
for image classification~\cite{LeeXGZT15,SzegedyLJSRAEVR15}, 
is also adopted for helping deep networks training
and improving the heatmap estimation quality, e.g.,~\cite{WeiRKS16, NewellYD16, TompsonGJLB15, BelagiannisZ17, ChenWPZYS17}.
The hourglass approach~\cite{NewellYD16}
and the convolutional pose machine approach~\cite{WeiRKS16}
process the intermediate heatmaps as the input or a part of the input
of the remaining subnetwork.

\vspace{.1cm}
\noindent\textbf{Our approach.}
Our network connects high-to-low subnetworks in parallel.
It maintains high-resolution representations
through the whole process for spatially precise heatmap estimation.
It generates reliable high-resolution representations 
through repeatedly fusing the representations
produced by the high-to-low subnetworks.
Our approach is different from
most existing works,
which need a separate low-to-high upsampling process
and aggregate low-level and high-level representations.
Our approach, without using intermediate heatmap supervision, is superior in keypoint detection accuracy and efficient in computation complexity and parameters.

There are related multi-scale networks for classification and segmentation~\cite{CaiFFV16, ChenPKMY18, XieT15, ZhaoSQWJ17, KanazawaSJ14, XuXZYZ14, SamyAESE18, SaxenaV16, HuangCLWMW17, ZhouHZ15, SamyAESE18, pohlen2017FRRN,FourureEFMT017}.
Our work is partially inspired by some of them~\cite{SaxenaV16, HuangCLWMW17, ZhouHZ15, SamyAESE18},
and there are clear differences making them not applicable to our problem.
Convolutional neural fabrics~\cite{SaxenaV16}
and 
interlinked CNN~\cite{ZhouHZ15}
fail to produce high-quality segmentation results
because of a lack of proper design on 
each subnetwork (depth, batch normalization) 
and multi-scale fusion.
The grid network~\cite{FourureEFMT017}, 
a combination of many weight-shared U-Nets,
consists of two separate fusion processes across multi-resolution representations:
on the first stage, information is only
sent from high resolution to low resolution;
on the second stage, information is only sent
from low resolution to high resolution,
and thus less competitive. 
Multi-scale densenets~\cite{HuangCLWMW17} does not target 
and cannot generate reliable 
high-resolution representations.

\section{Approach}
Human pose estimation, a.k.a. keypoint detection,
aims to detect the locations of $K$ keypoints or parts
(e.g., elbow, wrist, etc) from an image $\mathbf{I}$
of size $W \times H \times 3$.
The state-of-the-art methods
transform this problem
to estimating $K$ heatmaps of size $W^{'} \times H^{'}$,
$\{\mathbf{H}_1, \mathbf{H}_2, \dots, \mathbf{H}_K\}$, 
where each heatmap $\mathbf{H}_k$ indicates 
the location confidence of the $k$th keypoint.

We follow the widely-adopted pipeline~\cite{NewellYD16,XiaoWW18, ChenWPZYS17}
to predict human keypoints  
using a convolutional network, 
which is composed of
a stem consisting of two strided convolutions decreasing
the resolution,
a main body
outputting the feature maps with the same resolution 
as its input feature maps,
and a regressor estimating the heatmaps
where the keypoint positions are chosen
and transformed to the full resolution.
We focus on the design of the main body
and introduce our High-Resolution Net (HRNet) that is depicted in Figure~\ref{fig:trianglenet}.

\vspace{.1cm}
\noindent\textbf{Sequential multi-resolution subnetworks.}
Existing networks for pose estimation
are built by connecting high-to-low resolution subnetworks in series,
where each subnetwork, forming a stage, 
is composed of a sequence of convolutions
and there is a down-sample layer across adjacent subnetworks
to halve the resolution.

Let $\mathcal{N}_{sr}$ be the subnetwork
in the $s$th stage
and 
$r$ be the resolution index
(Its resolution is $\frac{1}{2^{r-1}}$ of the resolution
of the first subnetwork).
The high-to-low network with $S$ (e.g., $4$) stages can be denoted as:
\begin{equation}
\begin{array}{llll}
\mathcal{N}_{11}  & \rightarrow ~~ \mathcal{N}_{22} &\rightarrow ~~ \mathcal{N}_{33} & \rightarrow ~~ \mathcal{N}_{44}.
\end{array}
\end{equation}

\vspace{.1cm}
\noindent\textbf{Parallel multi-resolution subnetworks.}
We start from a high-resolution subnetwork as the first stage,
gradually add high-to-low resolution subnetworks one by one,
forming new stages,
and connect the multi-resolution subnetworks in parallel.
As a result,
the resolutions for the parallel subnetworks of a later stage  
consists of the resolutions from the previous stage,
and an extra lower one.

An example network structure, containing $4$ parallel subnetworks, 
is given as follows,
\begin{equation}
\begin{array}{llll}
\mathcal{N}_{11}  & \rightarrow ~~ \mathcal{N}_{21} &\rightarrow ~~ \mathcal{N}_{31} & \rightarrow ~~ \mathcal{N}_{41} \\
& \searrow ~~ \mathcal{N}_{22} &\rightarrow ~~ \mathcal{N}_{32} & \rightarrow ~~ \mathcal{N}_{42} \\
&   &\searrow  ~~ \mathcal{N}_{33} & \rightarrow ~~ \mathcal{N}_{43}  \\
&   & & \searrow ~~ \mathcal{N}_{44}. 
\end{array}
\end{equation}

\begin{figure}[t]
\small
\centering
\includegraphics[width=1.0\linewidth]{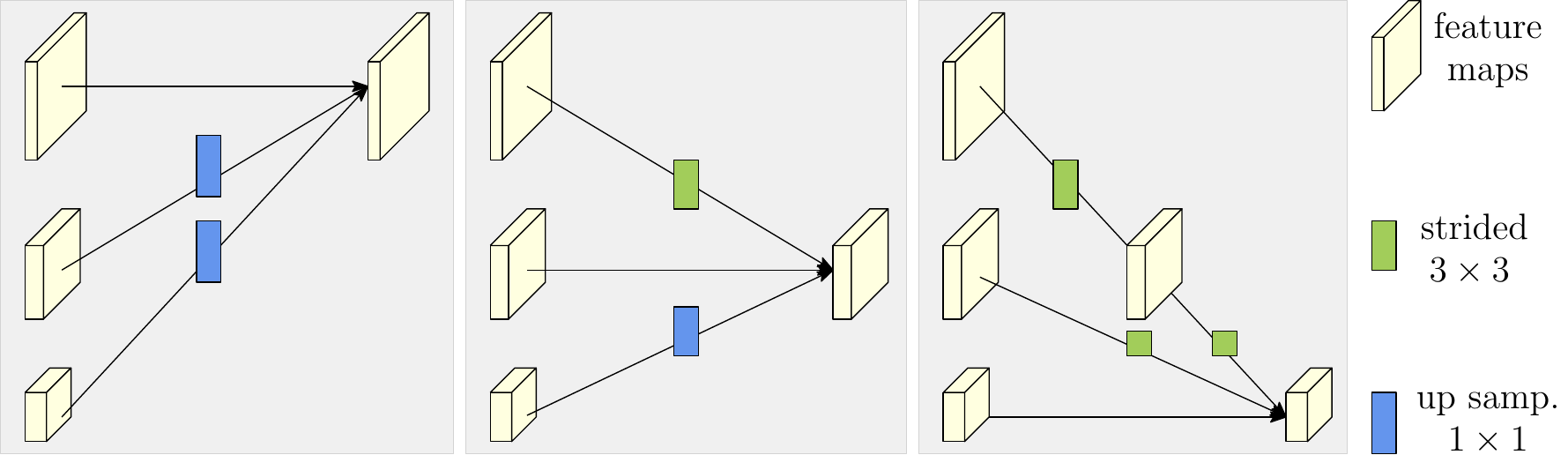}\\
\vspace{.2cm}
\caption{\small Illustrating
how 
the exchange unit
aggregates the information
for high, medium and low resolutions
from the left to the right,
respectively.
Right legend: strided $3\times 3$ = strided $3\times 3$ convolution,
up samp. $1\times 1$ = nearest neighbor up-sampling following a $1\times 1$ convolution.}
\label{fig:exchangeunit}
\end{figure}

\vspace{.1cm}
\noindent\textbf{Repeated multi-scale fusion.}
We introduce \emph{exchange units}
across parallel subnetworks
such that each subnetwork repeatedly receives
the information from other parallel subnetworks.
Here is an example showing the scheme
of exchanging information. 
We divided the third stage into several (e.g., $3$) exchange blocks,
and each block is composed of $3$ parallel convolution units
with an exchange unit across the parallel units, 
which
is given as follows,
\begin{equation}
\begin{array}{llllll}
\mathcal{C}^1_{31}  & \searrow &\nearrow ~~ \mathcal{C}^2_{31} & \searrow  &\nearrow ~~ \mathcal{C}^3_{31} & \searrow \\
\mathcal{C}^1_{32}  & \rightarrow ~~  \mathcal{E}^1_{3} &  \rightarrow ~~ \mathcal{C}^2_{32} & \rightarrow ~~  \mathcal{E}^2_{3} &  \rightarrow ~~ \mathcal{C}^3_{32} & \rightarrow ~~  \mathcal{E}^3_{3},\\
\mathcal{C}^1_{33}  &  \nearrow &\searrow ~~ \mathcal{C}^2_{33} &  \nearrow &\searrow ~~ \mathcal{C}^3_{33} &  \nearrow\\
\end{array}
\end{equation}
where $\mathcal{C}^b_{sr}$
represents the convolution unit in the $r$th resolution
of the $b$th block in the $s$th stage,
and $\mathcal{E}^b_{s}$ is the corresponding exchange unit.

We illustrate the exchange unit
in Figure~\ref{fig:exchangeunit}
and
present the formulation in the following.
We drop the subscript $s$ and the superscript $b$ 
for discussion convenience.
The inputs are $s$ response maps:
$\{\mathbf{X}_{1}, \mathbf{X}_{2}, \dots, \mathbf{X}_{s}\}$.
The outputs are $s$ response maps:
$\{\mathbf{Y}_{1}, \mathbf{Y}_{2}, \dots, \mathbf{Y}_{s}\}$,
whose resolutions and widths are the same to the input.
Each output is an aggregation of the input maps,
$\mathbf{Y}_{k}
= \sum_{i=1}^s a(\mathbf{X}_{i}, k)$.
The exchange unit across stages has an extra output map $\mathbf{Y}_{s+1}$: 
$\mathbf{Y}_{s+1}
= a(\mathbf{Y}_{s}, s+1)$.

The function $a(\mathbf{X}_{i}, k)$ 
consists of upsampling or downsampling 
$\mathbf{X}_{i}$ from resolution $i$ to resolution $k$.
We adopt strided $3 \times 3$ convolutions for downsampling.
For instance,
one strided $3 \times 3$ convolution with the stride $2$
for $2\times$ downsampling,
and two consecutive 
strided $3 \times 3$ convolutions with the stride $2$
for $4\times$ downsampling.
For upsampling,
we adopt the simple nearest neighbor sampling 
following a $1\times 1$ convolution
for aligning the number of channels.
If $i=k$, 
$a(\cdot , \cdot )$ is just an identify connection:
$a(\mathbf{X}_{i}, k) = \mathbf{X}_i$.

\vspace{.1cm}
\noindent\textbf{Heatmap estimation.}
We regress the heatmaps
simply from the high-resolution representations
output by the last exchange unit,
which empirically works well.
The loss function, defined as 
the mean squared error,
is applied for comparing
the predicted heatmaps
and the groundtruth heatmaps.
The groundtruth heatmpas are generated 
by applying $2$D Gaussian 
with standard deviation of $1$ pixel
centered on the grouptruth location
of each keypoint.

\vspace{.1cm}
\noindent\textbf{Network instantiation}. 
We instantiate the network
for keypoint heatmap estimation
by following the design rule of ResNet
to distribute the depth to each stage
and the number of channels to each resolution.

The main body, i.e.,
our HRNet,
contains four stages
with four parallel subnetworks,
whose the resolution is gradually decreased to a half
and accordingly the width (the number of channels)
is increased to the double.
The first stage contains $4$ residual units 
where each unit, the same to the ResNet-$50$,
is formed by a bottleneck with the width $64$,
and is followed by one $3\times3$ convolution
reducing the width of feature maps
to $C$.
The $2$nd, $3$rd, $4$th stages
contain $1$, $4$, $3$ exchange blocks, respectively.
One exchange block 
contains $4$ residual units where each unit contains two $3\times3$ convolutions in each resolution
and an exchange unit across resolutions.
In summary, there are totally $8$ exchange units,
i.e., $8$ multi-scale fusions are conducted.

In our experiments,
we study one small net and one big net:
HRNet-W$32$ and 
HRNet-W$48$,
where $32$ and $48$
represent the widths ($C$)
of the high-resolution subnetworks 
in last three stages, respectively.
The widths of other three parallel subnetworks
are $64, 128, 256$ for HRNet-W$32$,
and $96, 192, 384$ for HRNet-W$48$.

\section{Experiments}
\label{sec:experiments}

\renewcommand{\arraystretch}{1.3}
\begin{table*}[t]
\footnotesize
\caption[Caption for LOF]{Comparisons on the COCO validation set. 
Pretrain = pretrain the backbone on the ImageNet classification task. 
OHKM = online hard keypoints mining~\cite{ChenWPZYS17}.}
\centering
\label{table:coco_val}
\begin{tabular}{l|l|c|r|r|c|cccccc}
\hline
Method & Backbone &  Pretrain & Input size & \#Params & GFLOPs & 
$\operatorname{AP}$ & $\operatorname{AP}^{50}$ & $\operatorname{AP}^{75}$ & $\operatorname{AP}^{M}$ & $\operatorname{AP}^{L}$ & $\operatorname{AR}$ \\
\hline
$8$-stage Hourglass~\cite{NewellYD16} & $8$-stage Hourglass & N &  $256 \times 192$ & $25.1$M & $14.3$&
$66.9$&$-$&$-$&$-$&$-$&$-$\\ 
CPN~\cite{ChenWPZYS17}& ResNet-50 & Y & $256 \times 192$ & $27.0$M & $6.20$&
$68.6$&$-$&$-$&$-$&$-$&$-$\\ 
CPN + OHKM~\cite{ChenWPZYS17} & ResNet-50 & Y & $256 \times 192$ & $27.0$M & $6.20$&
$69.4$&$-$&$-$&$-$&$-$&$-$\\ 
SimpleBaseline~\cite{XiaoWW18} & ResNet-50 & Y & $256\times192$  &$34.0$M &$8.90$
&${70.4}$ & ${88.6}$&${78.3}$&${67.1}$&${77.2}$&${76.3}$\\
SimpleBaseline~\cite{XiaoWW18} & ResNet-101 & Y & $256\times192$  &$53.0$M &$12.4$
&${71.4}$ & ${89.3}$&${79.3}$&${68.1}$&${78.1}$&${77.1}$\\
SimpleBaseline~\cite{XiaoWW18} & ResNet-152  & Y & $256\times192$ &$68.6$M &$15.7$
&${72.0}$ & ${89.3}$&${79.8}$&${68.7}$&${78.9}$&${77.8}$\\
\hline
HRNet-W$32$ & HRNet-W$32$ & N & $256\times 192$&  $28.5$M & $7.10$ &
$73.4$&$89.5$&$80.7$&$70.2$&$80.1$&$78.9$  \\
HRNet-W$32$ & HRNet-W$32$& Y & $256\times 192$&  $28.5$M & $7.10$ &
$74.4$&$90.5$&$81.9$&$70.8$&$81.0$&$79.8$  \\
HRNet-W$48$ & HRNet-W$48$& Y &  $256\times 192$&  $63.6$M &$14.6$ &
$75.1$&$90.6$&$82.2$&$71.5$&$81.8$&$80.4$  \\
\hline
SimpleBaseline~\cite{XiaoWW18} & ResNet-152  & Y &  $384\times288$     &$68.6$M &$35.6$
&${74.3}$ & ${89.6}$&${81.1}$&${70.5}$&${79.7}$&${79.7}$\\
HRNet-W$32$ &HRNet-W$32$ & Y &  $384\times 288$&  $28.5$M &$16.0$ & $75.8$&$90.6$&${82.7}$&$71.9$&$82.8$&$81.0$  \\
HRNet-W$48$ & HRNet-W$48$& Y &  $384\times 288$&  $63.6$M &$32.9$ & $\textbf{76.3}$&$\textbf{90.8}$&$\textbf{82.9}$&$\textbf{72.3}$&$\textbf{83.4}$&$\textbf{81.2}$  \\
\hline
\end{tabular}
\end{table*}

	\begin{table*}[t]
		\caption{Comparisons on the COCO test-dev set.   
			\#Params and FLOPs are calculated 
            for the pose estimation network, and 
            those for human detection and keypoint grouping are not included.}
	\centering
			\label{table:coco_test_dev}
			\footnotesize
            \begin{tabular}{l|l|c|c|c|lllllc}
				\hline
				Method &Backbone& Input size & \#Params & GFLOPs&
				$\operatorname{AP}$ & $\operatorname{AP}^{50}$ & $\operatorname{AP}^{75}$ & $\operatorname{AP}^{M}$ & $\operatorname{AP}^{L}$ & $\operatorname{AR}$\\
				\hline
				\multicolumn{11}{c}{Bottom-up: keypoint detection and grouping}\\
				\hline
				OpenPose~\cite{CaoSWS17} &$\hfil-$& $-$ &$-$& $-$  
				&$61.8$ & $84.9$&$67.5$&$57.1$&$68.2$&$66.5$\\
				Associative Embedding~\cite{NewellHD17} & $\hfil-$ & $-$ &$-$& $-$
				&$65.5$ & $86.8$&$72.3$&$60.6$&$72.6$&$70.2$\\
				PersonLab~\cite{PapandreouZCGTK18} & $\hfil-$ & $-$ &$-$& $-$
				&$68.7$ & $89.0$&$75.4$&$64.1$&$75.5$&$75.4$\\
				MultiPoseNet~\cite{KocabasKA18} & $\hfil-$ & $-$ &$-$& $-$
				&$69.6$ & $86.3$&$76.6$&$65.0$&$76.3$&$73.5$\\
				\hline 
				\multicolumn{11}{c}{Top-down: human detection and single-person keypoint detection}\\
				\hline
				Mask-RCNN~\cite{HeGDG17} & ResNet-50-FPN& $-$ &$-$& $-$
				& $63.1$ & $87.3$&$68.7$&$57.8$&$71.4$&$-$\\
				G-RMI~\cite{PapandreouZKTTB17} & ResNet-101 & $353\times257$ &$42.6$M& $57.0$
				&$64.9$ & $85.5$&$71.3$&$62.3$&$70.0$&$69.7$\\
				Integral Pose Regression~\cite{SunXWLW18} & ResNet-101 & $256\times256$ &$45.0$M& $11.0$
				&$67.8$ & $88.2$&$74.8$&$63.9$&$74.0$&$-$\\
				G-RMI + extra data~\cite{PapandreouZKTTB17} & ResNet-101 & $353\times257$ &$42.6$M& $57.0$
				&$68.5$ & $87.1$&$75.5$&$65.8$&$73.3$&$73.3$\\
				CPN~\cite{ChenWPZYS17} & ResNet-Inception& $384\times288$ &$-$& $-$
				& $72.1$ & $91.4$&$80.0$&$68.7$&$77.2$&$78.5$\\
				RMPE~\cite{FangXTL17} & PyraNet~\cite{YangLOLW17} & $320\times256$ &$28.1$M& $26.7$
				&$72.3$ & $89.2$&$79.1$&$68.0$&$78.6$&$-$\\
				CFN~\cite{HuangGT17} & $\hfil-$ & $-$ &$-$& $-$
				& $72.6$ & $86.1$&$69.7$&$78.3$&$64.1$&$-$\\
				CPN (ensemble)~\cite{ChenWPZYS17} & ResNet-Inception& $384\times288$ &$-$& $-$
				&$73.0$ & $91.7$&$80.9$&$69.5$&$78.1$&$ 79.0$\\
				SimpleBaseline~\cite{XiaoWW18} & ResNet-152&$384\times288$  &$68.6$M& $35.6$
				&${73.7}$ & ${91.9}$&${81.1}$&${70.3}$&${80.0}$&${79.0}$\\
				\hline
				HRNet-W$32$ & HRNet-W$32$& $384\times 288$ &$28.5$M&$16.0$ &$74.9$&$92.5$&$82.8$&$71.3$&$80.9$&$80.1$\\
				HRNet-W$48$ & HRNet-W$48$& $384\times 288$ &$63.6$M& $32.9$
				& $\textbf{75.5}$&$\textbf{92.5}$&$\textbf{83.3}$&$\textbf{71.9}$&$\textbf{81.5}$&$\textbf{80.5}$\\
				\hline
				HRNet-W$48$ + extra data & HRNet-W$48$& $384\times 288$ &$63.6$M& $32.9$
				& $\textbf{77.0}$&$\textbf{92.7}$&$\textbf{84.5}$&$\textbf{73.4}$&$\textbf{83.1}$&$\textbf{82.0}$\\
				\hline
			\end{tabular}
	\end{table*}

\subsection{COCO Keypoint Detection}
\label{sec:coco}
\noindent\textbf{Dataset.}
The COCO dataset~\cite{LinMBHPRDZ14} contains over 
$200,000$ images and $250,000$ person instances labeled with $17$ keypoints. 
We train our model on COCO train$2017$ dataset, including $57K$ images 
and $150K$ person instances. 
We evaluate our approach on the val$2017$ set and test-dev$2017$ set, 
containing $5000$ images and $20K$ images, respectively.

\vspace{.1cm}
\noindent\textbf{Evaluation metric.}
The standard evaluation metric
is
based on Object Keypoint Similarity (OKS):
$\operatorname{OKS} = \frac{\sum_{i}\exp(-d_i^2/2s^2k_i^2)\delta(v_i > 0)}{\sum_i \delta(v_i > 0)}.$
Here $d_i$ is the Euclidean distance between 
the detected keypoint and the corresponding ground truth,
$v_i$ is the visibility flag of the ground truth,
$s$ is the object scale, and 
$k_i$ is a per-keypoint constant that controls falloff.
We report standard average precision and recall scores\footnote{\url{http://cocodataset.org/\#keypoints-eval}}:
$\operatorname{AP}^{50}$ ($\operatorname{AP}$ at $\operatorname{OKS} = 0.50$)
$\operatorname{AP}^{75}$,
$\operatorname{AP}$ 
(the mean of $\operatorname{AP}$ scores at $10$ positions,
$\operatorname{OKS} = 0.50, 0.55, \dots,0.90, 0.95$;
$\operatorname{AP}^M$  for medium objects,
$\operatorname{AP}^L$  for large objects,
and $\operatorname{AR}$ at $\operatorname{OKS} = 0.50, 0.55, \dots,0.90, 0.955$.

\vspace{.1cm}
\noindent\textbf{Training.}
We extend the human detection box in height or width 
to a fixed aspect ratio:
$\operatorname{height}: \operatorname{width} = 4 : 3$,
and then crop the box from the image,
which is resized to a fixed size, $256 \times 192$ or $384 \times 288$.
The data augmentation includes
random rotation ($[\ang{-45}, \ang{45}] $),
random scale ($[0.65, 1.35]$), and flipping. Following ~\cite{wang2018mscoco}, half body data augmentation is also involved.

We use the Adam optimizer~\cite{KingmaB14}.
The learning schedule
follows the setting~\cite{XiaoWW18}.
The base learning rate is set as $1\mathrm{e}{-3}$,
and is dropped to $1\mathrm{e}{-4}$ and $1\mathrm{e}{-5}$ 
at the $170$th and $200$th epochs, respectively.
The training process is terminated within $210$ epochs.

\vspace{.1cm}
\noindent\textbf{Testing.}
The two-stage top-down paradigm similar as~\cite{PapandreouZKTTB17,ChenWPZYS17,XiaoWW18}
is used:
detect the person instance using a person detector,
and then predict 
detection keypoints. 

We use the same person detectors provided 
by SimpleBaseline\footnote{\url{https://github.com/Microsoft/human-pose-estimation.pytorch}}~\cite{XiaoWW18} for both validation set and test-dev set. 
Following the common practice~\cite{XiaoWW18,NewellYD16,ChenWPZYS17},
we compute the heatmap by averaging the headmaps of the original and flipped images.
Each keypoint location is predicted
by adjusting the highest heatvalue location with a quarter offset
in the direction from the highest response
to the second highest response.

\vspace{.1cm}
\noindent\textbf{Results on the validation set.}
We report the results of our method and other state-of--the-art methods
in Table~\ref{table:coco_val}.
Our small network - HRNet-W$32$, trained from scratch
with the input size $256 \times 192$,
achieves an $73.4$ AP score,
outperforming other methods 
with the same input size.
(\romannum{1}) Compared to Hourglass~\cite{NewellYD16},
our small network improves AP by $6.5$ points, and the GFLOPs of our network is much lower and less than half,
while the number of parameters are similar 
and ours is slightly larger.
(\romannum{2}) Compared to CPN~\cite{ChenWPZYS17} w/o and w/ OHKM, our network, with slightly larger model size and slightly higher complexity, achieves $4.8$ and $4.0$ points gain, respectively.
(\romannum{3}) Compared to the previous best-performed SimpleBaseline~\cite{XiaoWW18},
our small net HRNet-W$32$
obtains significant improvements: $3.0$ points gain for the backbone ResNet-$50$ with a similar model size and GFLOPs, 
and $1.4$ points gain for the backbone ResNet-$152$ whose model size (\#Params) and GLOPs
are twice as many as ours.

Our nets can benefit from 
(\romannum{1}) training
from the model pretrained for the ImageNet classification problem:
The gain is $1.0$ points for HRNet-W$32$;
(\romannum{2}) increasing the capacity by increasing the width:
Our big net HRNet-W$48$ gets $0.7$ and $0.5$ 
improvements for the input sizes $256\times192$ and $384\times288$, respectively.

Considering the input size $384 \times 288$,
our HRNet-W$32$ and HRNet-W$48$,
get the $75.8$ and $76.3$ AP, which have $1.4$ and $1.2$ improvements
compared to the input size $256 \times 192$.
In comparison to the SimpleBaseline~\cite{XiaoWW18} that uses ResNet-$152$ as the backbone,
our HRNet-W$32$ and HRNet-W$48$ attain $1.5$ and $2.0$ points gain in terms of AP
at $45\%$ and $92.4\%$ computational cost, respectively.

\vspace{.1cm}
\noindent\textbf{Results on the test-dev set.}
Table~\ref{table:coco_test_dev} reports the pose estimation performances of our approach
and the existing state-of-the-art approaches.
Our approach is significantly better than 
bottom-up approaches.
On the other hand,
our small network, HRNet-W$32$, 
achieves an AP of $74.9$.
It outperforms all the other top-down approaches,
and is more efficient in terms of model size (\#Params) and computation complexity (GFLOPs).
Our big model, HRNet-W$48$, achieves the highest $75.5$ AP.
Compared to the SimpleBaseline~\cite{XiaoWW18}
with the same input size, 
our small and big networks receive $1.2$ and $1.8$ improvements, respectively.
With additional data from AI Challenger~\cite{wu2017ai} for training, our single big network can obtain an AP of $77.0$.

\subsection{MPII Human Pose Estimation}
\noindent\textbf{Dataset.}
The {MPII} Human Pose dataset~\cite{AndrilukaPGS14} 
consists of images
taken from a wide-range of real-world activities with full-body pose annotations.
There are around $25K$ images
with $40K$ subjects,
where there are $12K$ subjects for testing
and the remaining subjects for
the training set.
The {data augmentation}
and the {training strategy} are the same to MS COCO, 
except that the input size is cropped to $256\times256$ for fair comparison with other methods.

\vspace{.1cm}
\noindent\textbf{Testing.}
The testing procedure is almost the same to
that in COCO 
except that 
we adopt the standard testing strategy
to use the provided person boxes 
instead of detected person boxes. 
Following~\cite{ChuYOMYW17,YangLOLW17,TangYW18}, a six-scale pyramid testing 
procedure is performed. 

\vspace{.1cm}
\noindent\textbf{Evaluation metric.}
The standard metric~\cite{AndrilukaPGS14},
the PCKh (head-normalized probability of correct keypoint) score,
is used. 
A joint is correct if it falls
within $\alpha l$ pixels of the groundtruth position, 
where $\alpha$ is a constant and $l$ is the head size
that corresponds to $60\%$ of the diagonal length of
the ground-truth head bounding box.
The PCKh$@0.5$ ($\alpha=0.5$) score is reported.

\vspace{.1cm}
\noindent\textbf{Results on the test set.}
Tables~\ref{tab:mpii_test} and~\ref{tab:mpii_param_flops} 
show the PCKh$@0.5$ results, the model size and the GFLOPs of the top-performed methods.
We reimplement the SimpleBaseline~\cite{XiaoWW18} 
by using ResNet-$152$ as the backbone with the input size $256\times256$.
Our HRNet-W$32$ achieves a $92.3$ PKCh@$0.5$ score, and 
outperforms the stacked hourglass approach~\cite{NewellYD16}
and its extensions~\cite{SunLXZLW17,ChuYOMYW17,YangLOLW17,KeCQL18,TangYW18}.
Our result is the same 
as the best one~\cite{TangYW18}
among the previously-published results
on the leaderboard of Nov. $16$th, $2018$\footnote{\url{http://human-pose.mpi-inf.mpg.de/\#results}}.
We would like to point out that the approach~\cite{TangYW18},
complementary to our approach,
exploits the compositional model to learn the configuration of human bodies and 
adopts multi-level intermediate supervision, from which our approach 
can also benefit.
We also tested our big network - HRNet-W$48$ 
and obtained the same result $92.3$.
The reason might be that the performance in this datatset
tends to be saturate.

\begin{table}[t]
	\centering
	\tabcolsep=2.6pt    
	\footnotesize
    \centering
    \caption{Performance comparisons on the MPII test set (PCKh$@0.5$).}
	\label{tab:mpii_test}
	\begin{tabular}{l|ccccccc|cc}
		\hline%
		  Method& Hea &	Sho &	Elb	& Wri	& Hip	& Kne	& Ank &	Total \\
		\hline
        Insafutdinov et al.~\cite{InsafutdinovPAA16} & $96.8$ & $95.2$ & $89.3$ & $84.4$ & $88.4$ & $83.4$ & $78.0$ & $88.5$ \\
		Wei et al.~\cite{WeiRKS16} & $97.8$ & $95.0$ & $88.7$ & $84.0$ & $88.4$ & $82.8$ & $79.4$ & $88.5$ \\
        Bulat et al.~\cite{BulatT16} & $97.9$ & $95.1$ & $89.9$ & $85.3$ & $89.4$ & $85.7$ & $81.7$ & $89.7$ \\
		Newell et al.~\cite{NewellYD16} & $98.2$ & $96.3$ & $91.2$ & $87.1$ & $90.1$ & $87.4$ & $83.6$ & $90.9$ \\
		Sun et al.~\cite{SunLXZLW17}  & $98.1$ & $96.2$ & $91.2$ & $87.2$ & $89.8$ & $87.4$ & $84.1$ & $91.0$ \\
		Tang et al.~\cite{TangPGWZM18}  & $97.4$ & $96.4$ & $92.1$ & $87.7$ & $90.2$ & $87.7$ & $84.3$ & $91.2$ \\
        Ning et al.~\cite{NingZH18}  & $98.1$ & $96.3$ & $92.2$ & $87.8$ & $90.6$ & $87.6$ & $82.7$ & $91.2$ \\
		Luvizon et al.~\cite{Luvizon17} & $98.1$ & $96.6$ & $92.0$ & $87.5$ & $90.6$ & $88.0$ & $82.7$ & $91.2$ \\
		Chu et al.~\cite{ChuYOMYW17} & $98.5$ & $96.3$ & $91.9$ & $88.1$ & $90.6$ & $88.0$ & $85.0$ & $91.5$ \\
        Chou et al.~\cite{ChouCC17} & $98.2$ & $96.8$ & $92.2$ & $88.0$ & $91.3$ & $89.1$ & $84.9$ & $91.8$ \\
        Chen et al.~\cite{ChenSWLY17} & $98.1$ & $96.5$ & $92.5$ & $88.5$ & $90.2$ & $\textbf{89.6}$ & $86.0$ & $91.9$ \\
		Yang et al.~\cite{YangLOLW17} & $98.5$ & $96.7$ & $92.5$ & $88.7$ & $91.1$ & $88.6$ & $86.0$ & $92.0$ \\  
		Ke et al.~\cite{KeCQL18} & $98.5$ & $96.8$ & $92.7$ & $88.4$ & $90.6$ & $89.3$ & $\textbf{86.3}$ & $92.1$ \\  
		Tang et al.~\cite{TangYW18} & $98.4$ & $\textbf{96.9}$ & $92.6$ & $88.7$ & $\textbf{91.8}$ & $89.4$ & $86.2$ & $\textbf{92.3}$ \\ 
        \hline
        SimpleBaseline~\cite{XiaoWW18}& $98.5$ & $96.6$ & $91.9$ & $87.6$ & $91.1$ & $88.1$ & $84.1$ & $91.5$ \\
        HRNet-W$32$ &$\textbf{98.6}$  & $\textbf{96.9}$  & $\textbf{92.8}$  & $\textbf{89.0}$  & $91.5$  & $89.0$ & $85.7$ & $\textbf{92.3}$ \\		
		\hline
	\end{tabular}	
\end{table}

\renewcommand{\arraystretch}{1.3}
\begin{table}[bpt]
	\caption{\#Params and GFLOPs of some top-performed methods reported in Table~\ref{tab:mpii_test}.
    The GFLOPs is computed with 
    the input size $256\times256$.}
	\setlength{\tabcolsep}{18pt}
    \tabcolsep=3.6pt    
	\footnotesize
	\label{tab:mpii_param_flops}
	\centering
	\begin{tabular}{l|c|c|c}
		\hline
		Method & \#Params & GFLOPs &PCKh@$0.5$\\
		\hline
		Insafutdinov et al.~\cite{InsafutdinovPAA16}& $42.6$M & $41.2$&$88.5$\\
		Newell et al.~\cite{NewellYD16} & $25.1$M & $19.1$ &$90.9$\\
		Yang et al.~\cite{YangLOLW17} & $28.1$M & $21.3$  &$92.0$ \\
		Tang et al.~\cite{TangYW18} & $15.5$M & $15.6$   &$92.3$ \\
        \hline
		 SimpleBaseline~\cite{XiaoWW18}&$68.6$M&$20.9$ &$91.5$\\
		 HRNet-W$32$ &$28.5$M&$9.5$&$92.3$ \\
		\hline
	\end{tabular}
\end{table}

\subsection{Application to Pose Tracking}
\noindent\textbf{Dataset.}
PoseTrack~\cite{IqbalMG17} is a large-scale benchmark for human pose estimation and articulated tracking in video.
The dataset,
based on the raw videos provided by the popular MPII Human
Pose dataset,
contains $550$ video sequences
with $66,374$ frames.
The video sequences are split
into $292$, $50$, $208$
videos for training, validation, and testing, respectively.
The length of the training videos
ranges between $41-151$ frames,
and $30$ frames from the center of the video are densely annotated. 
The number
of frames in the validation/testing videos ranges between
$65-298$ frames. 
The $30$
frames around the keyframe from the MPII Pose dataset 
are densely annotated, and afterwards
every fourth frame is annotated.
In total, this constitutes roughly $23,000$
labeled frames and $153,615$ pose annotations.

\vspace{.1cm}
\noindent\textbf{Evaluation metric.}
We evaluate the results from two aspects: frame-wise
multi-person pose estimation, and multi-person pose tracking.
Pose estimation is evaluated by the mean Average Precision (mAP) as done in~\cite{PishchulinITAAG16, IqbalMG17}. Multi-person pose tracking is evaluated by the multi-object tracking accuracy (MOTA)~\cite{MilanL0RS16, IqbalMG17}.
Details are given in~\cite{IqbalMG17}.

\vspace{.1cm}
\noindent\textbf{Training.}
We train our HRNet-W$48$ for single person pose estimation
on the PoseTrack$2017$ training set,
where the network is initialized 
by the model
pre-trained on COCO dataset. 
We extract the person box, as the input of our network, 
from the annotated keypoints in the training frames by extending the bounding box
of all the keypoints (for one single person)
by $15\%$ in length.
The training setup, including data augmentation, 
is almost the same as that for COCO
except
that the learning schedule is different (as now it is for fine-tuning):
the learning rate starts from $1\mathrm{e}{-4}$,
drops to $1\mathrm{e}{-5}$ at the $10$th epoch,
and to $1\mathrm{e}{-6}$ at the $15$th epoch;
the iteration ends within $20$ epochs.

\vspace{.1cm}
\noindent\textbf{Testing.}
We follow~\cite{XiaoWW18}
to track poses across frames.
It consists of three steps:
person box detection and propagation,
human pose estimation,
and pose association cross nearby frames.
We use the same person box detector as used in SimpleBaseline~\cite{XiaoWW18},
and propagate the detected box into nearby frames
by propagating the predicted keypoints 
according to the optical flows computed
by FlowNet 2.0~\cite{IlgMSKDB17}\footnote{\url{https://github.com/NVIDIA/flownet2-pytorch}},
followed by
non-maximum suppression
for box removing.
The pose association scheme
is based on the object keypoint similarity
between the keypoints in one frame
and the keypoints propagated from the nearby frame
according to the optical flows.
The greedy matching algorithm
is then used
to compute the correspondence 
between keypoints in nearby frames.
More details are given in~\cite{XiaoWW18}.

\vspace{.1cm}
\noindent\textbf{Results on the PoseTrack$2017$ test set.}
Table \ref{table:posetrack_leaderboard} reports the results.
Our big network - HRNet-W$48$ 
achieves the superior result,
a $74.9$ mAP score and a $57.9$ MOTA score.
Compared with the second best approach,
the FlowTrack in SimpleBaseline~\cite{XiaoWW18},
that uses ResNet-$152$ as the backbone, 
our approach gets $0.3$ and $0.1$ points gain in terms of mAP and MOTA, respectively.
The superiority over the FlowTrack~\cite{XiaoWW18}
is consistent to 
that on the COCO keypoint detection
and MPII human pose estimation datasets.
This further implies the effectiveness of our pose estimation network.

\begin{table}[t]
\footnotesize
\begin{center}

\caption{Results of pose tracking on
the PoseTrack$2017$ test set.}
\label{table:posetrack_leaderboard}
\begin{tabular}{l|l|cc}
\hline
Entry &Additional training Data &mAP&MOTA\\
\hline
ML-LAB~\cite{zhu2017multi}&COCO+MPII-Pose&$70.3$&$41.8$\\
SOPT-PT~\cite{posetrackleaderboard}&COCO+MPII-Pose&$58.2$&$42.0$\\
BUTD2~\cite{jin2017towards}&COCO&$59.2$&$50.6$\\
MVIG~\cite{posetrackleaderboard}&COCO+MPII-Pose&$63.2$&$50.7$\\
PoseFlow~\cite{posetrackleaderboard}&COCO+MPII-Pose&$63.0$&$51.0$\\
ProTracker~\cite{girdhar2018detect}&COCO&$59.6$&$51.8$\\
HMPT~\cite{posetrackleaderboard}&COCO+MPII-Pose& $63.7$ & $51.9$ \\
JointFlow~\cite{doering2018joint}&COCO& $63.6$ & $53.1$ \\
STAF~\cite{posetrackleaderboard}&COCO+MPII-Pose& $70.3$ & $53.8$ \\
MIPAL~\cite{posetrackleaderboard}&COCO& $68.8$ & $54.5$ \\
FlowTrack~\cite{XiaoWW18}&COCO& $74.6$ & $57.8$ \\
\hline
HRNet-W$48$&COCO&$\textbf{74.9}$&$\textbf{57.9}$\\
\hline
\end{tabular}
\end{center}
\vspace{-3mm}
\end{table}

\subsection{Ablation Study}
We study the effect
of each component in our approach
on the COCO keypoint detection dataset. 
All results are obtained 
over the input size of $256\times192$
except the study 
about the effect of the input size.

\renewcommand{\arraystretch}{1.5}
\begin{table}[bpt]
\setlength{\tabcolsep}{3pt}
\scriptsize
\caption{Ablation study of exchange units that are used
in repeated multi-scale fusion.
Int. exchange across = intermediate exchange across stages,
Int. exchange within = intermediate exchange within stages.}
\label{tab:ablation_exchange_units}
\centering
	\begin{tabular}{c|c|c|c|c}
	\hline
	Method & Final exchange & Int. exchange across & Int. exchange within &  $\operatorname{AP}$ \\
	\hline
	(a) & \checkmark &  &    & $70.8$ \\
	(b) & \checkmark & \checkmark &    & $71.9$ \\
	(c) & \checkmark & \checkmark & \checkmark    & $73.4$ \\

	\hline
	\end{tabular}
\end{table}

\begin{figure*}[t]
	\centering
	\includegraphics[height = 0.129\textwidth]{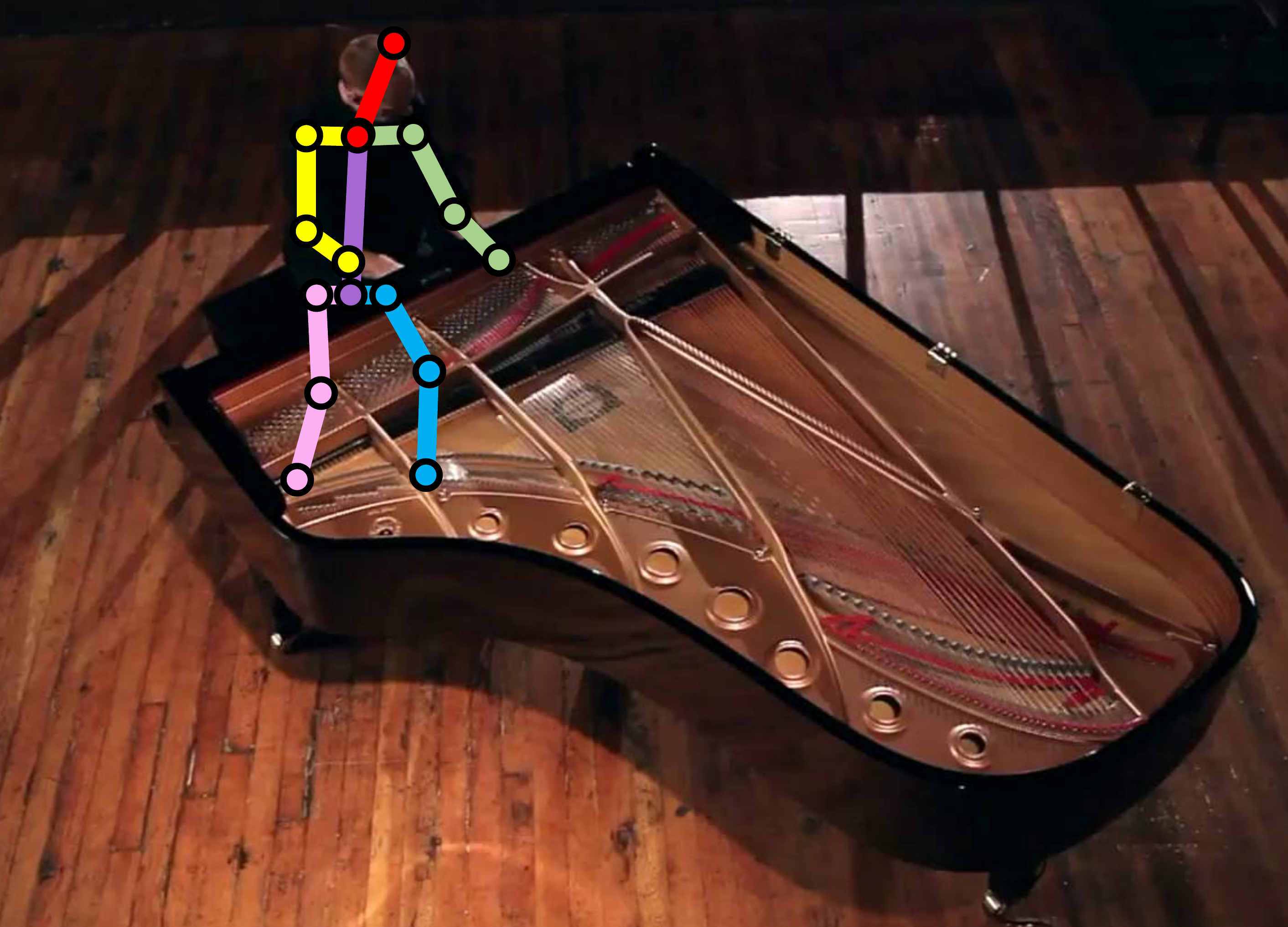}
	\includegraphics[height = 0.13\textwidth]{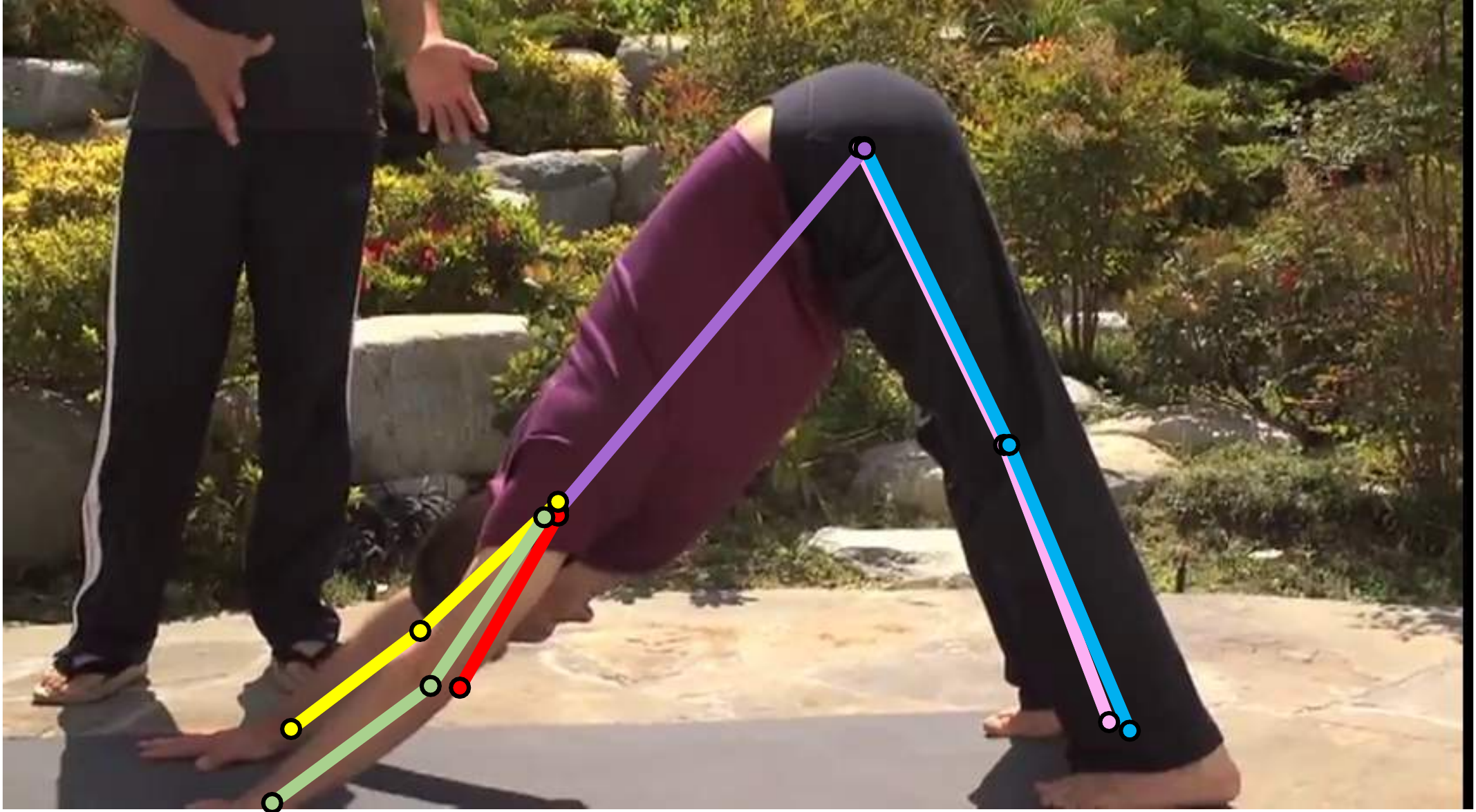}
	\includegraphics[height = 0.13\textwidth]{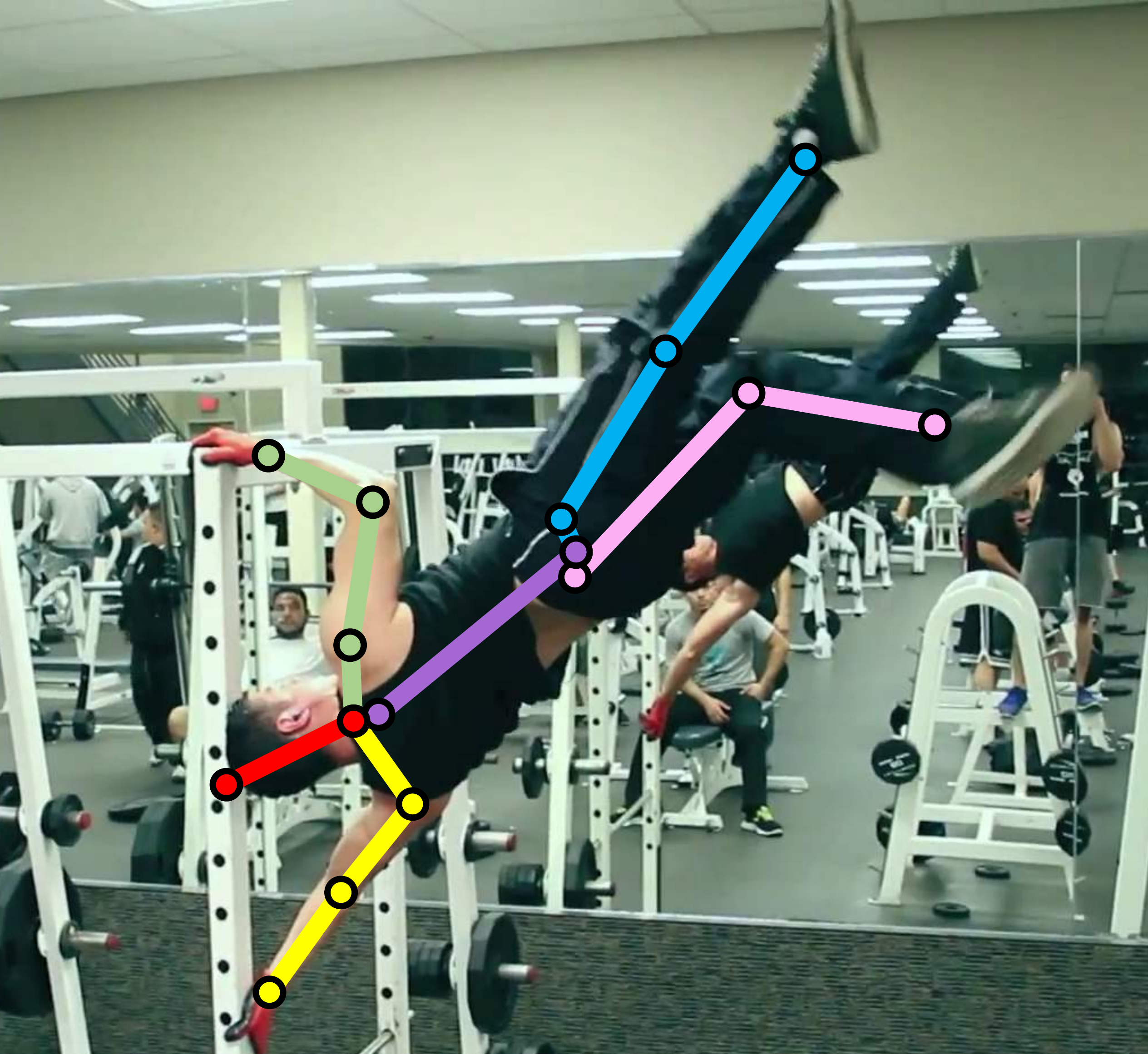}
	\includegraphics[height = 0.13\textwidth]{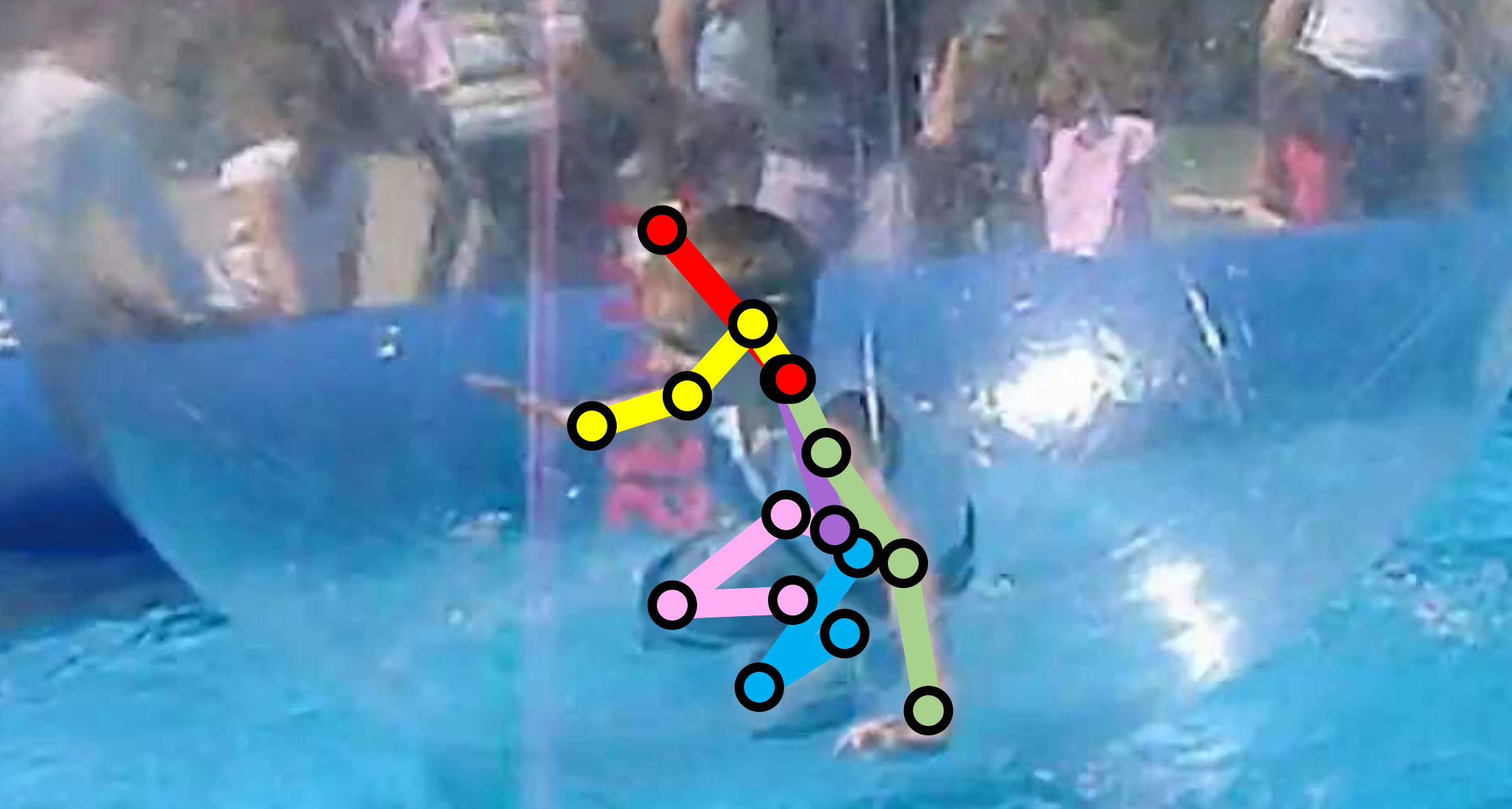}
	\includegraphics[height = 0.129\textwidth]{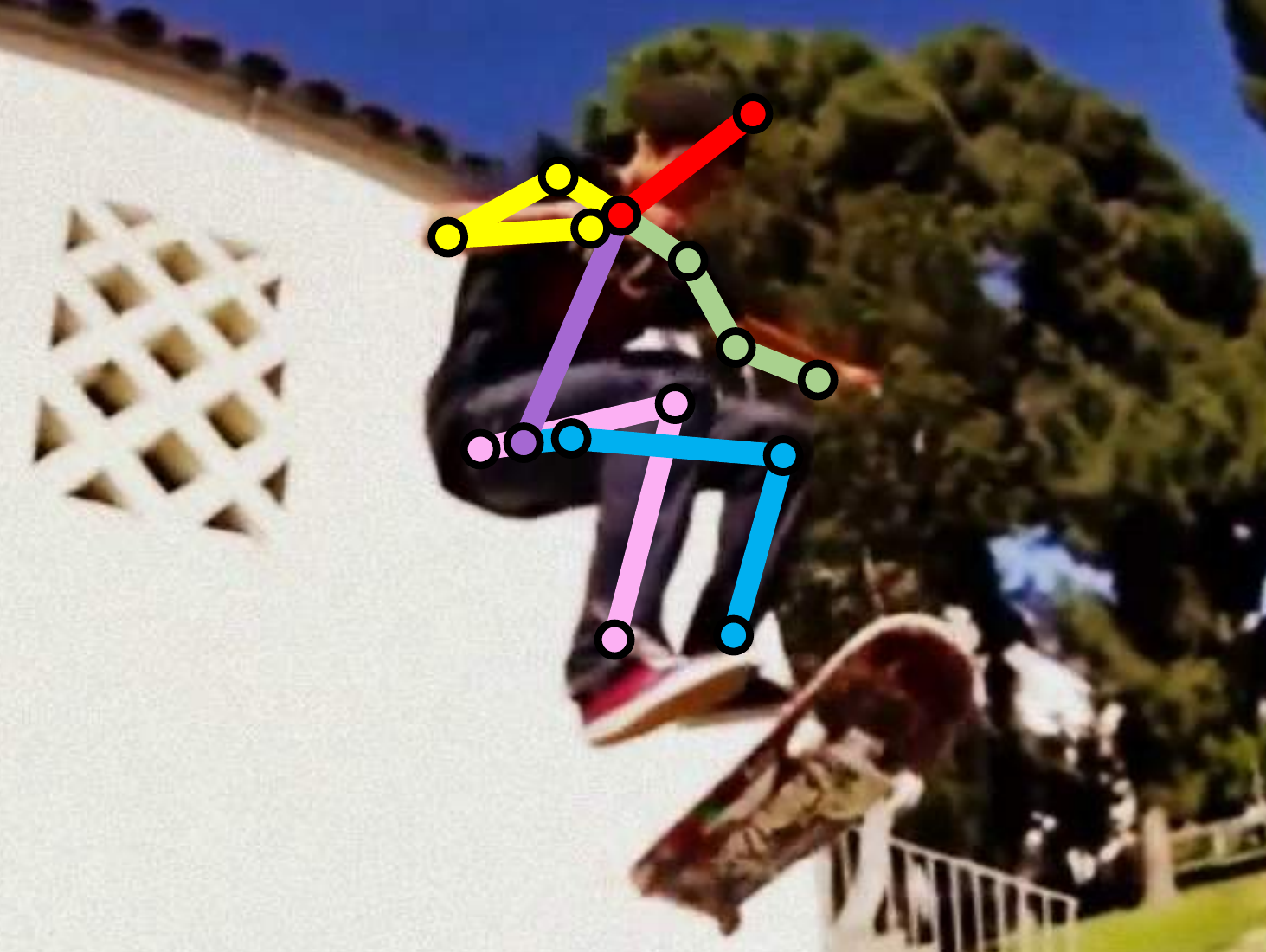}\\
	\includegraphics[height = 0.162\textwidth]{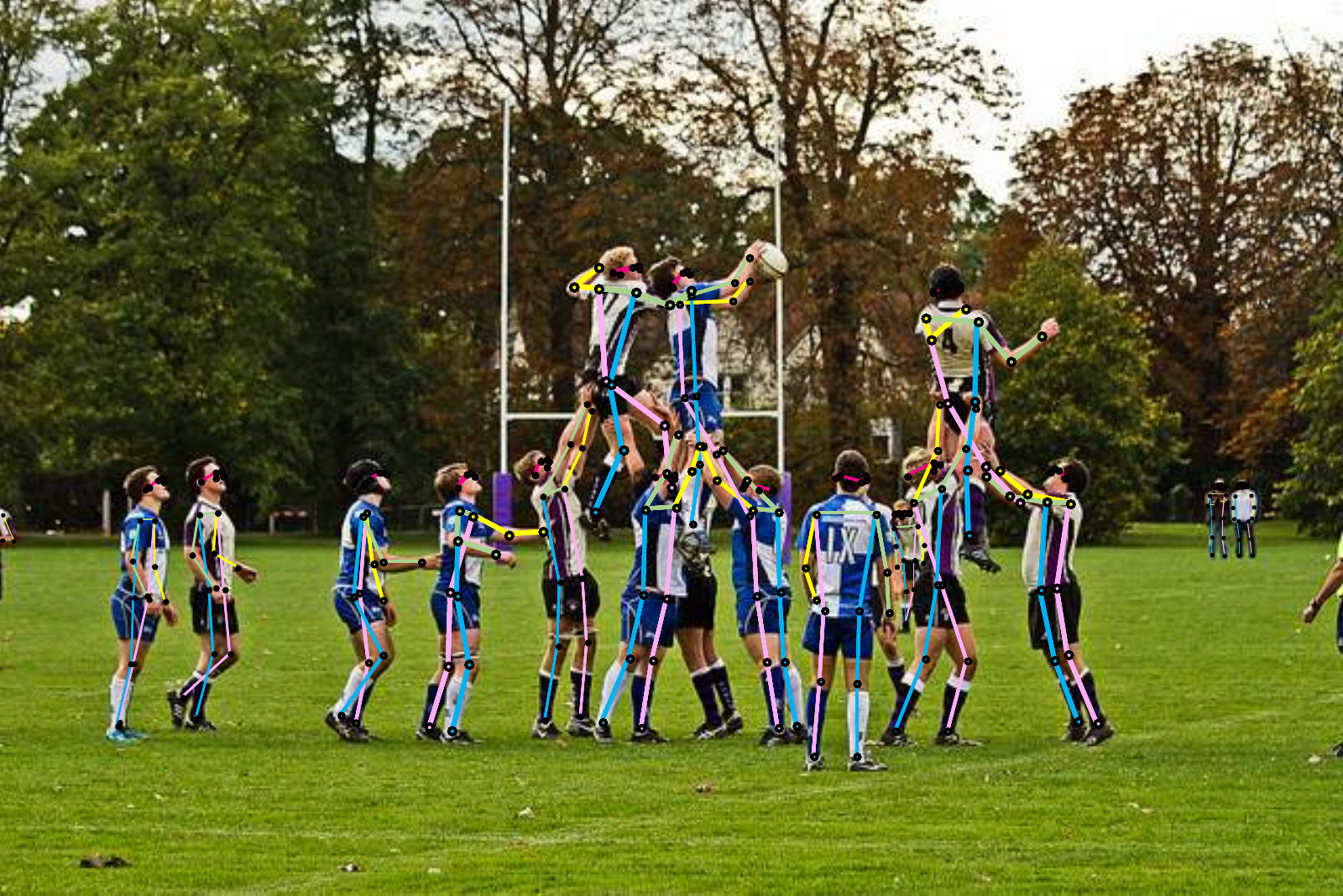}
	\includegraphics[height = 0.162\textwidth]{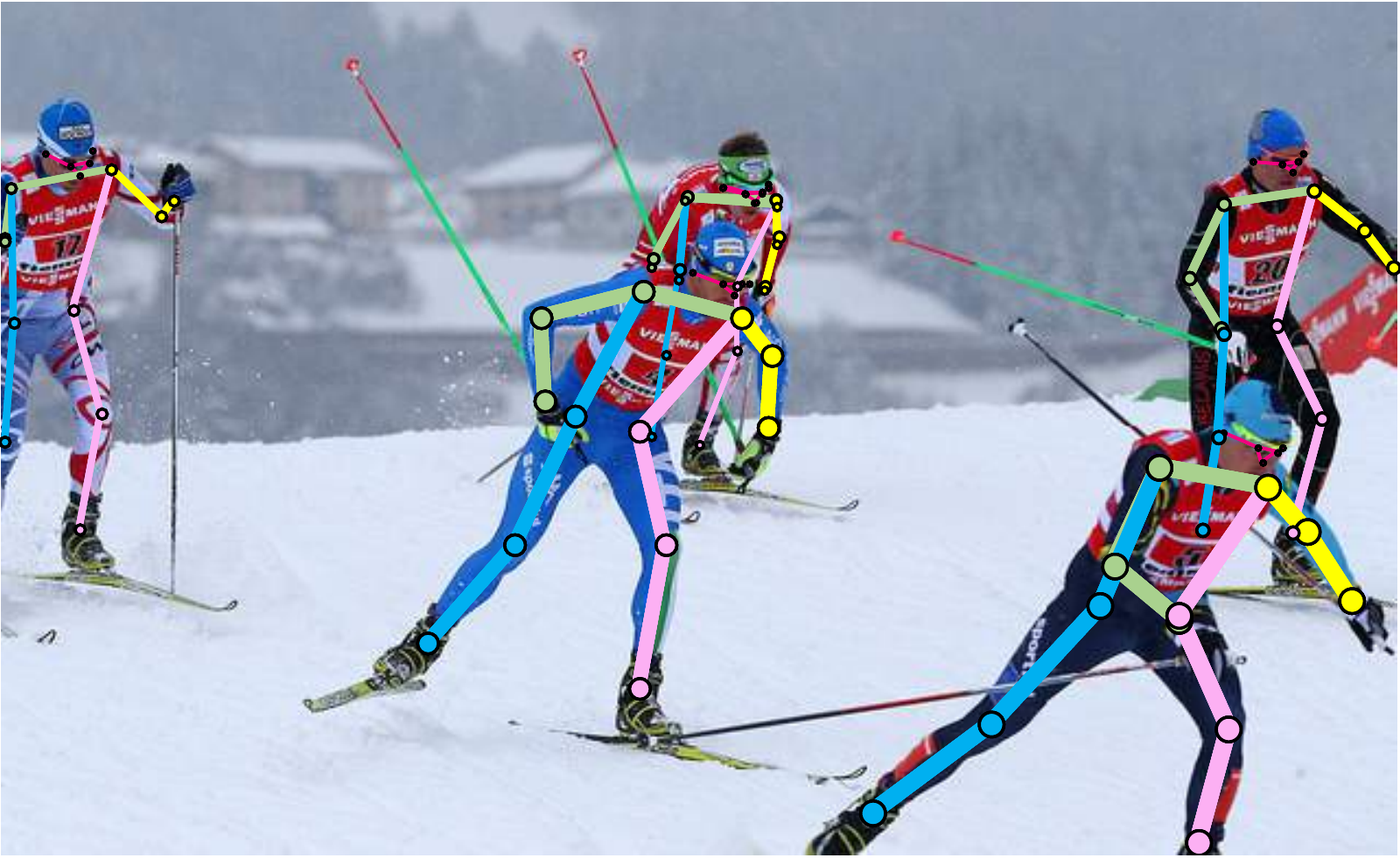}
	\includegraphics[height = 0.162\textwidth]{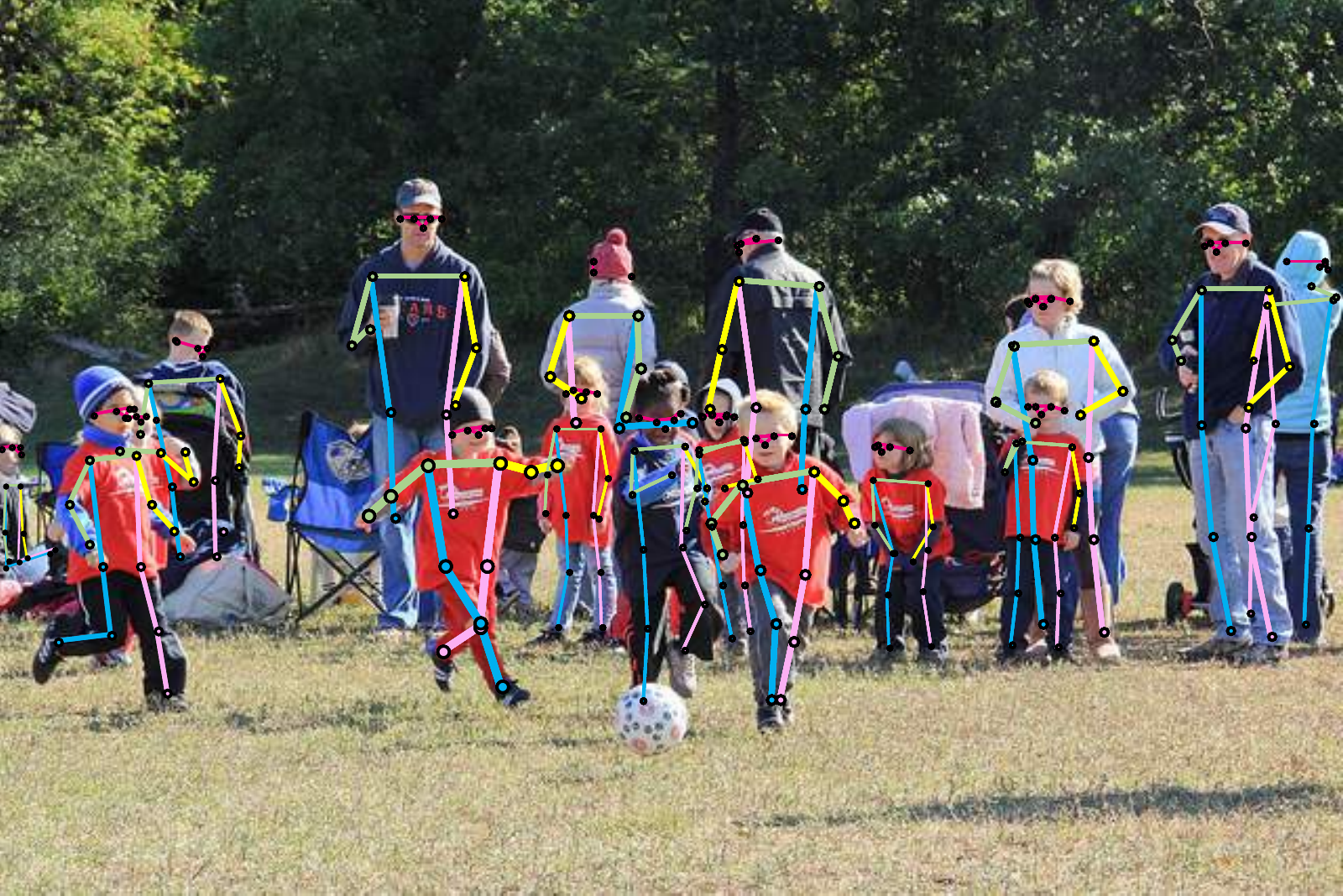}
	\includegraphics[height = 0.162\textwidth]{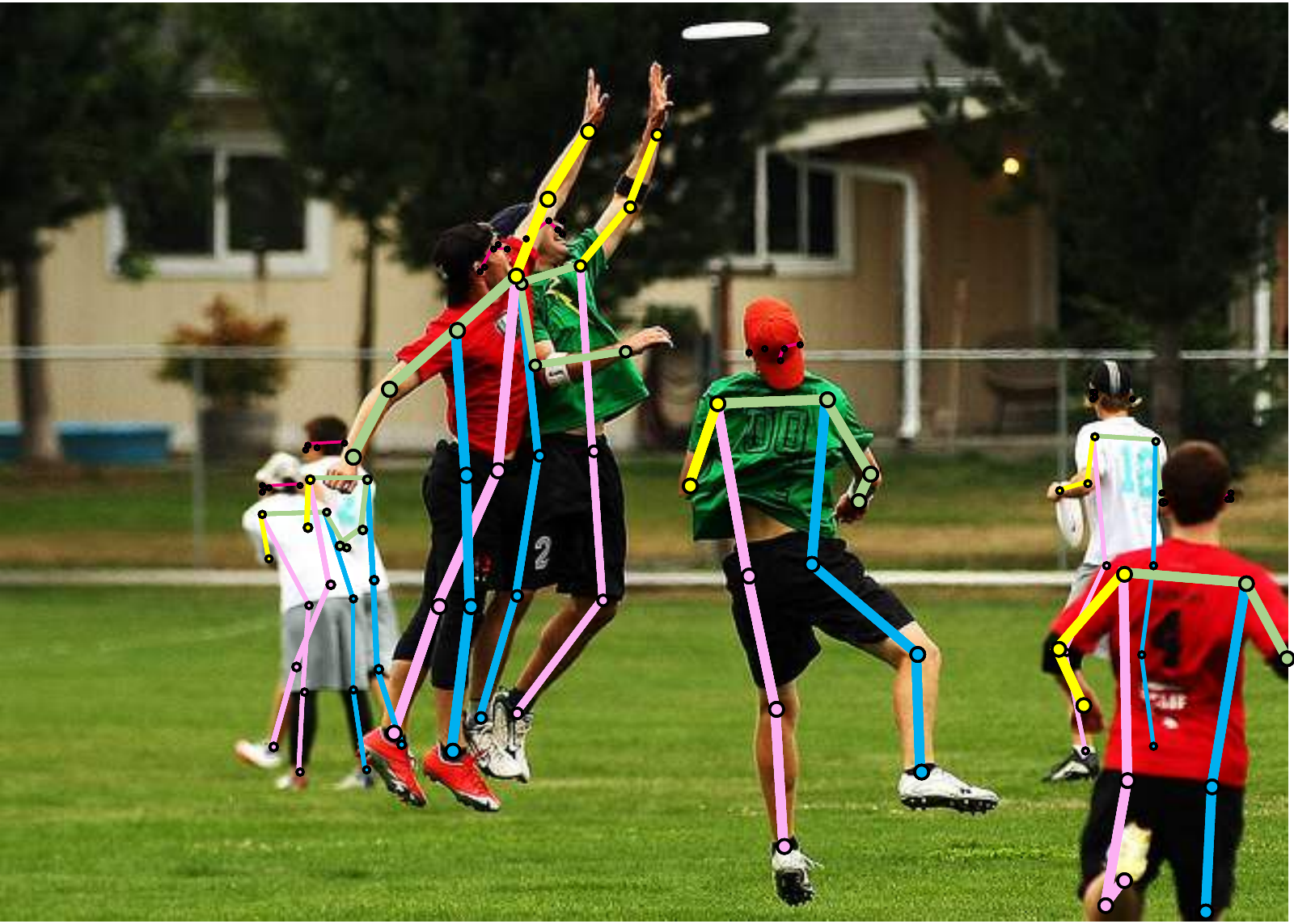}\\
	\caption{Qualitative results of some example images in the MPII (top) and COCO (bottom) datasets: containing viewpoint and appearance change, occlusion, multiple persons, and common imaging artifacts.}
\end{figure*}

\begin{figure}[t]
	\centering
	\includegraphics[width=0.9\linewidth]{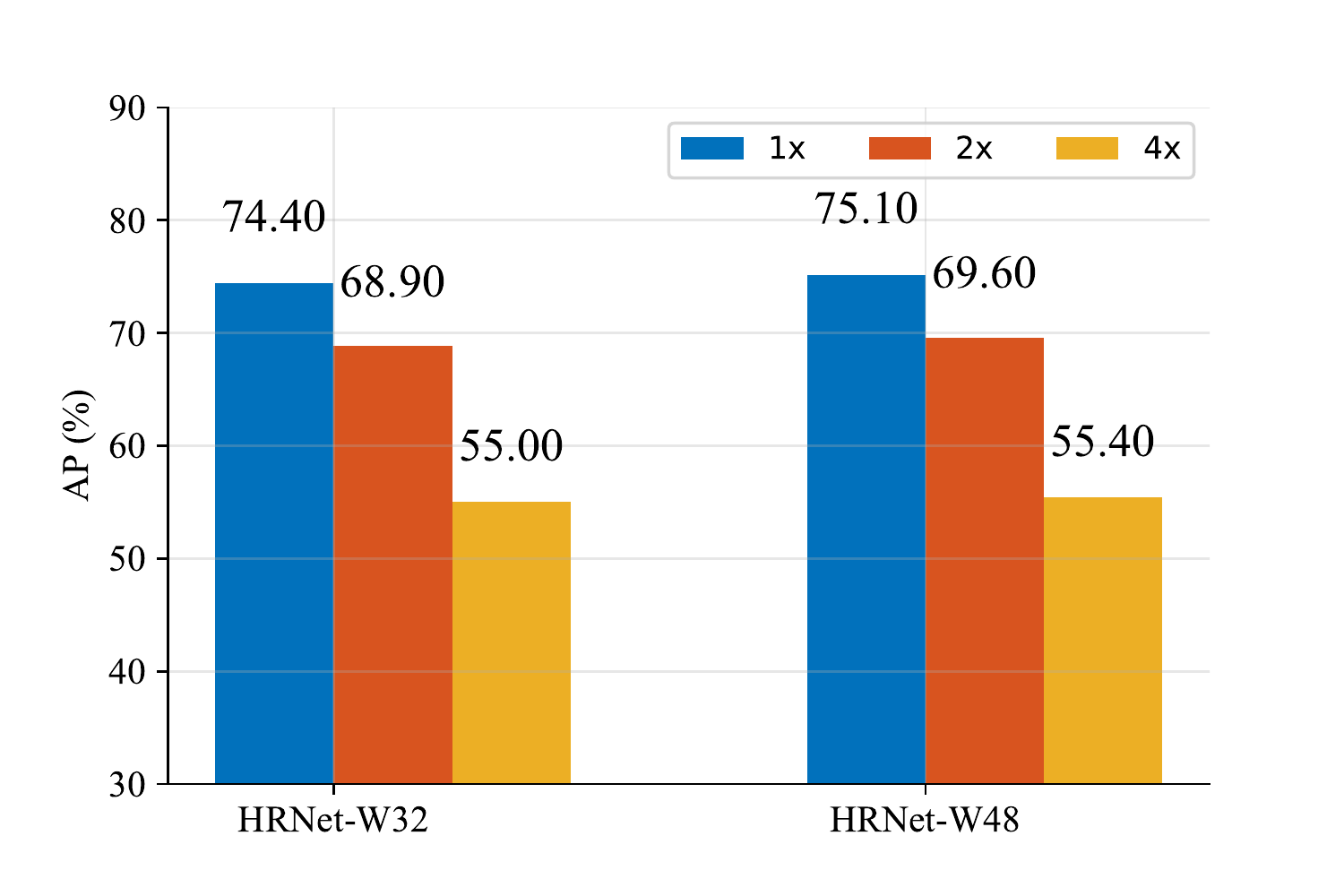}
	\caption{Ablation study of high and low representations.
			$1\times$, $2\times$, $4\times$ correspond to 
            the representations of 
            the high, medium, low resolutions, respectively. 
	}
	\label{fig:outputsize}
    \vspace{-3mm}
\end{figure}

\vspace{.1cm}
\noindent\textbf{Repeated multi-scale fusion.}
We empirically analyze 
the effect of the repeated multi-scale fusion.
We study three variants of our network.
(a) W/o intermediate exchange units ($1$ fusion):
There is no exchange between multi-resolution subnetworks
except the last exchange unit.
(b) W/ across-stage exchange units only ($3$ fusions):
There is no exchange between parallel subnetworks
within each stage.
(c) W/ both across-stage and within-stage exchange units 
(totally $8$ fusion):
This is our proposed method.
All the networks 
are trained from scratch.
The results on the COCO validation set given in Table~\ref{tab:ablation_exchange_units}
show that the multi-scale fusion is helpful
and more fusions lead to better performance.

\vspace{.1cm}
\noindent\textbf{Resolution maintenance.}
We study the performance of
a variant of the HRNet:
all the four high-to-low resolution subnetworks are added at the beginning
and the depth are the same; 
the fusion schemes are the same to ours.
Both our HRNet-W$32$ and the variant
(with similar \#Params and GFLOPs)
are trained from scratch
and tested on the COCO validation set.
The variant achieves an AP of $72.5$, which is lower than the $73.4$ AP of our small net, HRNet-W$32$.
We believe that the reason is that
the low-level features extracted from the early stages
over the low-resolution subnetworks
are less helpful.
In addition, 
the simple high-resolution network 
of similar parameter and computation complexities
without low-resolution parallel subnetworks
shows much lower performance .

\vspace{.1cm}
\noindent\textbf{Representation resolution.}
We study how the representation resolution affects the pose estimation performance 
from two aspects: 
check the quality of the heatmap estimated 
from the feature maps of
each resolution from high to low,
and study how the input size
affects the quality.

We train our small and big networks initialized 
by the model pretrained for the ImageNet classification.
Our network outputs four response maps from high-to-low solutions.
The quality of heatmap prediction over the lowest-resolution response map
is too low and the AP score is below $10$ points.
The AP scores over the other three maps
are reported in Figure~\ref{fig:outputsize}.
The comparison implies that 
the resolution does impact the keypoint prediction quality.

	\begin{figure}[t]
	\centering
	\includegraphics[width=0.9\linewidth]{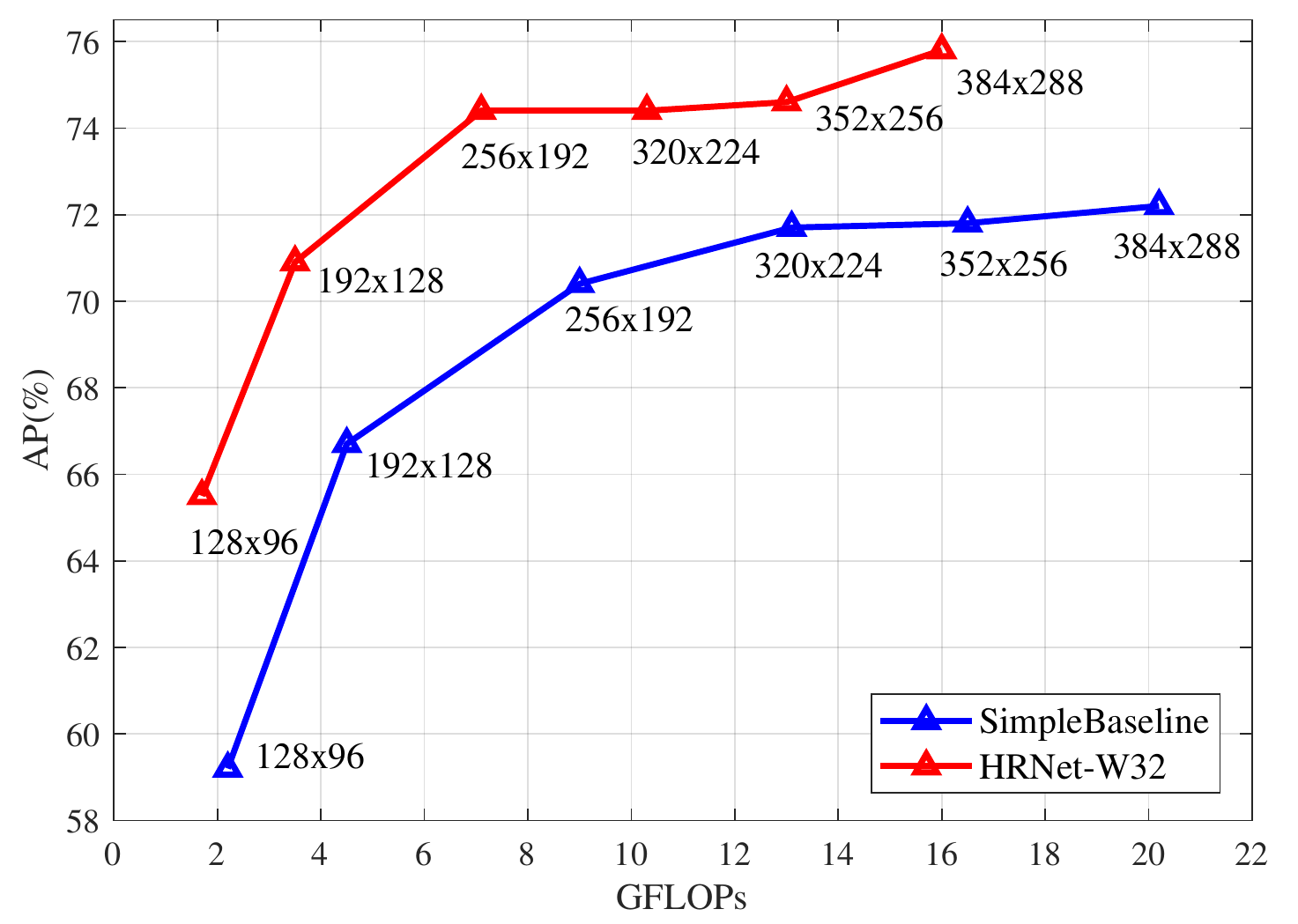}
	\caption{Illustrating how the performances of our HRNet 
    and SimpleBaseline~\cite{XiaoWW18} are affected by the input size.}
	\label{fig:inpsize_AP}
	\vspace{-2.5mm}
	\end{figure}

Figure~\ref{fig:inpsize_AP} shows how the input image size affects the performance in comparison with SimpleBaseline (ResNet-50)~\cite{XiaoWW18}.
We can find that the improvement for the smaller input size
is more significant
than the larger input size,
e.g., the improvement is $4.0$ points for $256\times192$
and $6.3$ points for $128\times96$.
The reason is that we maintain the high resolution 
through the whole process.
This implies that our approach is more advantageous
in the real applications where the computation cost 
is also an important factor.
On the other hand,
our approach with the input size $256\times192$ 
outperforms the SimpleBaseline~\cite{XiaoWW18} with the large input size of $384\times288$.

\section{Conclusion and Future Works}
In this paper,
we present a high-resolution network for human pose estimation,
yielding accurate and spatially-precise keypoint heatmaps.
The success stems from two aspects:
(\romannum{1}) maintain the high resolution through the whole process
without the need of recovering the high resolution;
and (\romannum{2}) fuse multi-resolution representations repeatedly,
rendering reliable high-resolution representations.

The future works include the applications
to other dense prediction tasks,
e.g., semantic segmentation, object detection,
face alignment,
image translation, 
as well as the investigation
on aggregating
multi-resolution representations in a less light way.
All them are
available at
\url{https://jingdongwang2017.github.io/Projects/HRNet/index.html}.

\section*{Appendix}
	\subsection*{Results on the MPII Validation Set}
	We provide the results on the MPII validation set~\cite{AndrilukaPGS14}. Our models are trained on a subset of MPII training set and evaluate on a heldout validation set of 2975 images.
	The training procedure is the same to that for training on the whole MPII training set.
	The heatmap is computed as the average of the heatmaps of the original and flipped images for testing.
	Following~\cite{YangLOLW17, TangYW18}, we also perform six-scale pyramid testing procedure (multi-scale testing). The results are shown in Table~\ref{tab:mpii_valid}.
	\begin{table}[b]
		\centering
		\tabcolsep=3.2pt    
		\footnotesize
		\centering
		\caption{Performance comparisons on the MPII validation set (PCKh$@0.5$).}
		\label{tab:mpii_valid}
		\begin{tabular}{l|cccccccg}
			\hline%
			Method& Hea &	Sho &	Elb	& Wri	& Hip	& Kne	& Ank &	Total \\
			\hline
			\multicolumn{9}{c}{Single-scale testing}\\
			\hline
			Newell et al.~\cite{NewellYD16} & $96.5$ & $96.0$ & $90.3$ & $85.4$ & $88.8$ & $85.0$ & $81.9$ & $89.2$ \\
			Yang et al.~\cite{YangLOLW17} & ${96.8}$ & ${96.0}$ & $90.4$ & $86.0$ & $89.5$ & $85.2$ & $82.3$ & $89.6$ \\  
			Tang et al.~\cite{TangYW18} & $95.6$ & $95.9$ & ${90.7}$ & ${86.5}$ & ${89.9}$ & ${86.6}$ & ${82.5}$ & $89.8$ \\  
			\hline
			SimpleBaseline~\cite{XiaoWW18}& $97.0$ &  $95.9$ &  $90.3$ &  $85.0$ & $89.2$  & $85.3$ &$81.3$  & $89.6$ \\
			HRNet-W$32$ &$97.1$  & $95.9$  & $90.3$  & ${86.4}$  & $89.1$  & $87.1$ & ${83.3}$ & $\textbf{90.3}$ \\		
			\hline
			\multicolumn{9}{c}{Multi-scale testing}\\
			\hline
			Newell et al.~\cite{NewellYD16}& $97.1$ & $96.1$ & $90.8$ & $86.2$ & $89.9$ & $85.9$ & $83.5$ & $90.0$ \\
			Yang et al.~\cite{YangLOLW17} & $97.4$ & ${96.2}$ & ${91.1}$ & $86.9$ & $90.1$ & $86.0$ & $83.9$ & $90.3$ \\  
			Tang et al.~\cite{TangYW18} & $97.4$ & ${96.2}$ & $91.0$ & $86.9$ & ${90.6}$ & $86.8$ & ${84.5}$ & $90.5$ \\  
			\hline
			SimpleBaseline~\cite{XiaoWW18}&  $97.5$ &  $96.1$ & $90.5$  & $85.4$  &  $90.1$ & $85.7$ & $82.3$ & $90.1$ \\
			HRNet-W$32$ &${97.7}$  & $96.3$  & $90.9$  &${86.7}$  & $89.7$  & ${87.4}$ & $84.1$ & $\textbf{90.8}$ \\
			\hline
		\end{tabular}
	\end{table}
	
	\subsection*{More Results on the PoseTrack Dataset}
	We provide the results for all the keypoints on the PoseTrack dataset~\cite{andriluka2018posetrack}. 
	Table~\ref{table:posetrack_pose_result} shows the multi-person pose estimation performance on the PoseTrack$2017$ dataset.
	Our HRNet-W$48$ achieves 77.3 and 74.9 points mAP on the validation and test setss, and outperforms previous state-of-the-art method~\cite{XiaoWW18} by 0.6 points and 0.3 points respectively.
	We provide more detailed results of multi-person pose tracking performance on the PoseTrack2017 test set as a supplement of the results reported in the paper, shown in Table~\ref{table:posetrack_tracking_result}.
	
	\setlength{\tabcolsep}{2.8pt}
	\begin{table}[t]
		\footnotesize
		\centering
		\caption{Multi-person pose estimation performance (MAP) on the PoseTrack2017 dataset. ``*'' means models trained on thr train+valid set. }%
		\label{table:posetrack_pose_result}
		\begin{tabular}{l|cccccccg}
			\hline 
			Method & Head 
			&  Sho. 
			& Elb. 
			& Wri.
			& Hip
			& Knee 
			& Ank.
			&Total\\
			\hline
			\multicolumn{9}{c}{PoseTrack validation set}\\
			\hline
			Girdhar et al.~\cite{girdhar2018detect} &$67.5$&$70.2$&$62.0$&$51.7$&$60.7$&$58.7$&$49.8$&$60.6$\\
			Xiu et al.~\cite{XiuLWFL18} &$66.7$&$73.3$&$68.3$&$61.1$&$67.5$&$67.0$&$61.3$&$66.5$\\
			Bin et al.~\cite{XiaoWW18} & $81.7$ & $83.4$ & $80.0$ & $72.4$ & $75.3$ & $74.8$ & $67.1$ & ${76.7}$ \\
			\hline
			HRNet-W$48$ & ${82.1}$ & ${83.6}$ & ${80.4}$ & ${73.3}$ & ${75.5}$ & ${75.3}$ & ${68.5}$ & $\textbf{77.3}$ \\
			\hline
			\multicolumn{9}{c}{PoseTrack test set}\\
			\hline
			Girdhar et al.*~\cite{girdhar2018detect} & $-$&$-$&$-$&$-$&$-$&$-$&$-$&$59.6$\\
			Xiu et al.~\cite{XiuLWFL18} &  $64.9$&$67.5$&$65.0$&$59.0$&$62.5$&$62.8$&$57.9$&$63.0$\\
			Bin et al.*~\cite{XiaoWW18} & ${80.1}$ & ${80.2}$ & ${76.9}$ & $71.5$ & $72.5$ & $72.4$& $65.7$ & ${74.6}$ \\
			\hline
			HRNet-W$48$*&${80.1}$ & ${80.2}$ & ${76.9}$ & ${72.0}$ & ${73.4}$ & ${72.5}$ & ${67.0}$ & $\textbf{74.9}$ \\
			\hline
		\end{tabular}
	\end{table}
	
	\setlength{\tabcolsep}{3pt}
	\begin{table}[t]
		\footnotesize
		\centering
			\caption{Multi-person pose tracking performance (MOTA) on the PoseTrack2017 test set.``*'' means models trained on the train+validation set.}%
			\label{table:posetrack_tracking_result}
			\begin{tabular}{l|cccccccg}
				\hline
				Method &  Head &   Sho. &  Elb. & Wri & Hip &  Knee &  Ank. & Total \\
				\hline
				Girdhar et al.*~\cite{girdhar2018detect} & $-$&$-$&$-$&$-$&$-$&$-$&$-$&$51.8$\\
				Xiu et al.~\cite{XiuLWFL18} &$52.0$&$57.4$& $52.8$ &$46.6$&$51.0$&$51.2$&$45.3$&$51.0$ \\
				Xiao et al.*~\cite{XiaoWW18} & $67.3$ & $68.5$ & $52.3$ & $49.3$ & $56.8$ & $57.2$ & $48.6$ & ${57.8}$ \\
				\hline
				HRNet-W$48$* & $67.1$ & $68.9$ & $52.2$ & $49.6$ & $57.7$ & $57.0$ & $48.5$ & $\textbf{57.9}$ \\
				\hline
			\end{tabular}
	\end{table}
	
	\subsection*{Results on the ImageNet Validation Set}
	We apply our networks to image classification task. The models are trained and evaluated on the ImageNet 2013 classification dataset~\cite{RussakovskyDSKS15}. We train our models for 100 epochs with a batch size of 256. The initial learning rate is set to 0.1 and is reduced by 10 times at epoch 30, 60 and 90. 
	Our models can achieve comparable performance as those networks specifically designed for image classification, such as ResNet~\cite{HeZRS16}. Our HRNet-W$32$ has a single-model top-5 validation error of 6.5\% and has a single-model top-1 validation error of 22.7\% with the single-crop testing. Our HRNet-W$48$ gets better performance: 6.1\% top-5 errors and 22.1\% top-1 error. We use the models trained on the ImageNet dataset to initialize the parameters of our pose estimation networks.
	
\pagebreak
\noindent\textbf{Acknowledgements.} The authors thank Dianqi Li and Lei Zhang for helpful discussions.
{\small
\bibliographystyle{ieee}

}

\end{document}